\documentclass{article}

\usepackage{PRIMEarxiv}
\usepackage[T1]{fontenc}
\usepackage[utf8]{inputenc} 
\usepackage{amsmath}        
\usepackage{amssymb}        
\usepackage{amsfonts}       
\usepackage{nicefrac}       
\usepackage{microtype}      
\usepackage{graphicx}       
\graphicspath{{media/}}     
\usepackage{booktabs}       
\usepackage{multirow}
\usepackage{array}
\usepackage{tabularx}
\usepackage{caption}
\usepackage{subcaption}
\usepackage{float}
\usepackage{xcolor}         
\usepackage{hyperref}       
\hypersetup{hidelinks}
\usepackage{url}            
\usepackage{glossaries}     

\DeclareMathOperator*{\argmax}{arg\,max}

\newacronym{ap}{AP}{Average Precision}
\newacronym{audioprotopnet}{AudioProtoPNet}{Audio Prototypical Part Network}
\newacronym{auroc}{AUROC}{Area Under the Receiver Operating Characteristic Curve}
\newacronym{at}{AT}{adversarial training}
\newacronym{ate}{AT-E}{embedding-space adversarial training}
\newacronym{ato}{AT-O}{output-space adversarial training}
\newacronym{awp}{AWP}{Adversarial Weight Perturbation}
\newacronym{cmap}{cmAP}{Class Mean Average Precision}
\newacronym{cnn}{CNN}{Convolutional Neural Network}
\newacronym{dl}{DL}{Deep Learning}
\newacronym{drs}{DRS}{Deformation Robustness Score}
\newacronym{fft}{FFT}{Fast Fourier Transform}
\newacronym{fgsm}{FGSM}{Fast Gradient Sign Method}
\newacronym{ml}{ML}{Machine Learning}
\newacronym{ood}{OOD}{out-of-distribution}
\newacronym{ot}{OT}{ordinary training}
\newacronym{pgd}{PGD}{Projected Gradient Descent}
\newacronym{ppnet}{ProtoPNet}{Prototypical Part Network}
\newacronym{prs}{PRS}{Performance Robustness Score}
\newacronym{relu}{ReLU}{Rectified Linear Unit}
\newacronym{ssl}{SSL}{Self-Supervised Learning}
\newacronym{stft}{STFT}{Short-Time Fourier Transform}
\newacronym{tars}{TARS}{Total Adversarial Robustness Score}
\newacronym{trades}{TRADES}{TRadeoff-inspired Adversarial DEfense via Surrogate-loss minimization}
\newacronym{pam}{PAM}{Passive Acoustic Monitoring}
\newacronym{pow}{POW}{Powdermill Nature Reserve}

\newcolumntype{M}[1]{>{\centering\arraybackslash}m{#1}}
\newcolumntype{Y}{>{\centering\arraybackslash}X}

\newcommand{\maintextfloatbreak}{}
\newenvironment{appendixstricttable}
  {\begin{table}[!ht]}
  {\end{table}}
\newenvironment{tmlrappendixstricttable}
  {\begin{table}[!ht]}
  {\end{table}}
\newenvironment{appendixcompactstricttable}
  {\begin{table}[!ht]}
  {\end{table}}
\newenvironment{arxivcompactappendixtable}
  {\table[!ht]}
  {\endtable}
\newenvironment{appendixstrictfigure}
  {\begin{figure}[!ht]}
  {\end{figure}}

\newcommand{\PaperTitle}{Adversarial Training Improves Generalization Under Distribution Shifts in Bird Sound Classification}
\newcommand{\PaperAuthorRene}{Ren\'e Heinrich}
\newcommand{\PaperAuthorLukas}{Lukas Rauch}
\newcommand{\PaperAuthorRaphael}{Raphael Schwinger}
\newcommand{\PaperAuthorKatharina}{Katharina Brauns}
\newcommand{\PaperAuthorBastian}{Bastian Sch\"afermeier}
\newcommand{\PaperAuthorBernhard}{Bernhard Sick}
\newcommand{\PaperAuthorChristoph}{Christoph Scholz}
\newcommand{\PaperAffiliationIEE}{Fraunhofer Institute for Energy Economics and Energy System Technology (IEE), Kassel, Germany}
\newcommand{\PaperAffiliationIES}{Intelligent Embedded Systems (IES), University of Kassel, Kassel, Germany}
\newcommand{\PaperAffiliationINS}{Intelligent Systems (INS), Kiel University, Kiel, Germany}
\newcommand{\PaperCorrespondingEmail}{rene.heinrich@iee.fraunhofer.de}

\usepackage{etoolbox}
\renewcommand{\maintextfloatbreak}{\newpage}
\renewenvironment{appendixstricttable}
  {\begin{table}[H]\captionsetup{skip=5pt}}
  {\end{table}}
\renewenvironment{appendixcompactstricttable}
  {\begin{table}[H]\captionsetup{skip=3pt}}
  {\end{table}}
\renewenvironment{arxivcompactappendixtable}
  {\table[H]\captionsetup{font=small,skip=3pt}}
  {\endtable}
\renewenvironment{appendixstrictfigure}
  {\begin{figure}[H]\captionsetup{skip=5pt}}
  {\end{figure}}
\pretocmd{\appendix}{\raggedbottom}{}{}

\let\citet\cite
\let\citep\cite

\title{\PaperTitle}
\author{%
  \textbf{\PaperAuthorRene}\textsuperscript{1,2,*},
  \textbf{\PaperAuthorLukas}\textsuperscript{2},
  \textbf{\PaperAuthorRaphael}\textsuperscript{3}, \\
  \textbf{\PaperAuthorKatharina}\textsuperscript{1,2},
  \textbf{\PaperAuthorBastian}\textsuperscript{1},
  \textbf{\PaperAuthorBernhard}\textsuperscript{2},
  \textbf{\PaperAuthorChristoph}\textsuperscript{1,2} \\ \\
  \textsuperscript{1} \PaperAffiliationIEE \\
  Joseph-Beuys-Stra{\ss}e 8, 34117 Kassel, Germany \\ \\
  \textsuperscript{2} \PaperAffiliationIES \\
  M\"onchebergstra{\ss}e 19, 34127 Kassel, Germany \\ \\
  \textsuperscript{3} \PaperAffiliationINS \\
  Christian-Albrechts-Platz 4, 24118 Kiel, Germany \\ \\
  * \PaperCorrespondingEmail
}

\newcommand{\PaperAuthorContributions}{%
  \section*{CRediT authorship contribution statement}
  \noindent\textbf{\PaperAuthorRene:} Conceptualization, Methodology, Software, Validation, Formal analysis, Investigation, Data curation, Visualization, Writing -- original draft, Writing -- review \& editing.\par
  \noindent\textbf{\PaperAuthorLukas:} Writing -- original draft (related-work section and literature review), Writing -- review \& editing.\par
  \noindent\textbf{\PaperAuthorRaphael:} Software (implementation of the Perch-based distribution-shift quantification), Formal analysis (distribution-shift quantification), Writing -- original draft (distribution-shift quantification section), Writing -- review \& editing.\par
  \noindent\textbf{\PaperAuthorKatharina:} Funding acquisition, Writing -- review \& editing.\par
  \noindent\textbf{\PaperAuthorBastian:} Funding acquisition, Writing -- review \& editing.\par
  \noindent\textbf{\PaperAuthorBernhard:} Funding acquisition, Supervision, Writing -- review \& editing.\par
  \noindent\textbf{\PaperAuthorChristoph:} Funding acquisition, Project administration, Supervision, Writing -- review \& editing.
}

\newcommand{\PaperAcknowledgements}{%
  \PaperAuthorContributions
  \section*{Acknowledgements}
  This work was supported by several research initiatives.
  The investigation of prototype stability of \gls*{audioprotopnet} under targeted embedding-space attacks was conducted within the DeepBirdDetect project (Fkz. 67KI31040A), funded by the German Federal Ministry for the Environment, Nature Conservation and Nuclear Safety (BMUV).
  The development of untargeted embedding-space attacks and embedding-space adversarial training, together with the analysis of their effects on generalization and adversarial robustness, was carried out within the BioActiveAI project (Fkz. 24\_0081\_2B), funded by the Hessian Ministry for Digitalization and Innovation.
  The development of untargeted output-space attacks and output-space adversarial training, as well as the corresponding evaluation of generalization and empirical robustness under \gls*{pgd} attacks, was performed within the BioDroneAI project (Fkz. 02WDG1758A), funded by the German Federal Ministry of Research, Technology and Space (BMFTR).
}
\newcommand{\PaperAIDisclosure}{%
  \section*{Use of AI tools}
  The authors used OpenAI ChatGPT (GPT-5.6 Sol) and Anthropic Claude (Claude
Opus 5) as writing and coding assistants. All AI-suggested text and code were
reviewed, edited, and verified by the authors. The assistants did not conduct
model training or generate experimental results. The authors take full
responsibility for the resulting text and code.
}
\newcommand{\PaperBibliography}{%
  \bibliographystyle{unsrt}%
  \bibliography{audioprotopnet_arxiv}%
}

\begin{document}
\maketitle

\begin{abstract}
Adversarial training is a promising strategy for enhancing robustness against adversarial attacks, but its impact on generalization under substantial distribution shifts in audio classification remains largely unexplored.
We address this gap by investigating how adversarial training strategies improve generalization performance and adversarial robustness in audio classification.
We study two architectures for multi-label bird sound classification: ConvNeXt, a strong \gls*{cnn} baseline, and AudioProtoPNet, a prototype-based model that has demonstrated state-of-the-art performance while providing inherent interpretability through learned prototypes.
Experiments use BirdSet, a challenging benchmark for bird sound classification in bioacoustics.
Bioacoustic recordings exhibit substantial covariate shift due to heterogeneous recording devices and acoustic environments.
We compare adversarial training based on output-space attacks, which maximize classification loss, and embedding-space attacks, which maximize embedding dissimilarity.
Both attack types are also used for robustness evaluation.
Additionally, for AudioProtoPNet, the study assesses the stability of its learned prototypes under targeted embedding-space attacks. 
Results show that adversarial training, particularly using output-space attacks, improves clean test-data performance by up to 10.5\% relative in mean cmAP and simultaneously strengthens the adversarial robustness of the models.
These findings, although derived from the bird sound domain, suggest that adversarial training holds potential to enhance robustness against both strong distribution shifts and adversarial attacks in challenging audio classification settings.
\end{abstract}


\section{Introduction}
\label{sec:introduction}
Deep learning models often struggle to generalize from controlled training environments to real-world deployment, particularly in audio classification because of highly variable acoustic signals.
Audio classification models are vulnerable to distribution shifts that arise from heterogeneous recording devices (e.g., different microphone types), fluctuating noise (e.g., wind or traffic), and diverse acoustic environments (e.g., urban vs. forest settings) \cite{abesser2020review,abesser2021towards,krishnamachari2023mitigating}. 
Consequently, despite deep learning's progress across many audio tasks, state-of-the-art models exhibit performance drops under these unseen conditions in acoustic scene classification, speech recognition, and industrial anomaly detection \cite{kosmider2021spectrum,sanabria2023measuring,albertini2024imad}, leaving a persistent gap between benchmark performance and real-world efficacy \cite{kheddar2024automatic,krishnamachari2023mitigating,van2024birds}.

The problem is especially severe in bioacoustics, where animal vocalizations are influenced by habitat, sensor location, weather, and even regional dialects within the same species \cite{lostanlen2019robust,goerlitz2018weather,ghani2023global}. 
Consequently, bioacoustic data are often characterized by a range of distribution shifts, including covariate shift \cite{shimodaira2000improving,sugiyama2012machine,abesser2020review}, domain shift \cite{ben2010theory,ganin2015unsupervised}, task shift \cite{sun2024metalearning}, and subpopulation shift \cite{tran2022plex,yang2023change}. 
These complexities are further compounded by label uncertainty \cite{frenay2013classification,natarajan2013learning,fonseca2021fsd50k} and class imbalance \cite{nitesh2002smote,he2009learning}. 
Bird sound classification serves as an ideal case study for these challenges, as it involves a vast number of species recorded across diverse geographic regions and acoustic conditions \cite{rauch2024birdset}. 
In passive acoustic monitoring, soundscape recordings frequently contain overlapping vocalizations from multiple species, which naturally induces a multi-label classification setting at evaluation time.
Similar issues pervade other audio classification tasks, underscoring the need for robust generalization across application scenarios.

Traditional data augmentation techniques, such as Mixup \cite{zhang2017mixup, hendrycks2019augmix} and time-frequency masking \cite{park2019specaugment}, improve generalization but generally do not encompass the full variation encountered in real-world environments \cite{abayomi2022data,kumar2024improving,rauch2024birdset}.
Modern audio classification models typically rely on high-dimensional embeddings to represent complex acoustic features. 
Consequently, there is a need for training methodologies that promote the learning of more generalizable embeddings to prepare models for these data distribution shifts. 

\begin{figure}[!ht]
    \centering
    \includegraphics[trim={0cm 0cm 0cm 0cm}, clip, scale=0.48]{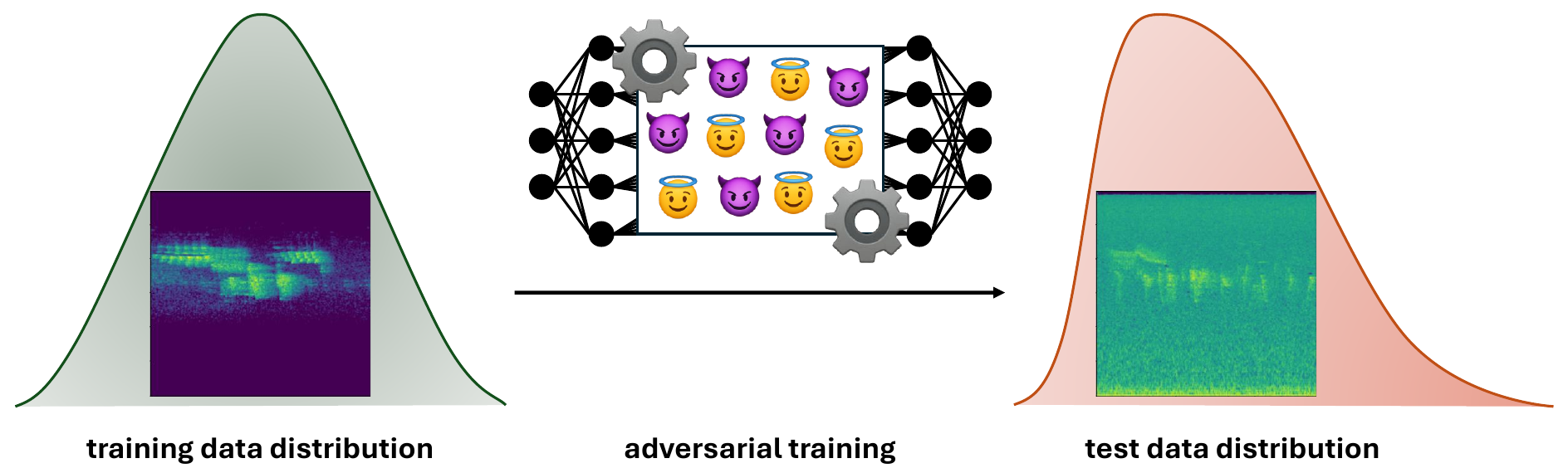}
    \caption{ Conceptual illustration of employing adversarial training to mitigate the effects of data distribution shifts between the training data distribution (left) and the test data distribution (right).
    }
\label{fig:illustration}
\end{figure}

Adversarial training \cite{madry2018towards} offers a promising approach to these limitations by augmenting training with perturbations specifically optimized to fool the network into making incorrect classifications, rather than with the random variations introduced by standard augmentation.
By explicitly learning to withstand these perturbations, models can develop embeddings that are inherently more robust to subtle changes in input signals. 
While initially conceived as a defense against malicious adversarial attacks, the core principle of adversarial training (i.e., forcing the model to learn invariant embeddings) holds potential as an augmentation technique to improve generalization under data distribution shifts. 
As illustrated in Figure~\ref{fig:illustration}, this training paradigm may enhance performance even on clean, unperturbed data when such shifts are prevalent.
We distinguish two complementary objectives: \emph{generalization}, defined as maintaining classification performance on clean but distributionally shifted test data, and \emph{adversarial robustness}, defined as maintaining performance under intentional, norm-bounded input perturbations.
Both objectives are relevant for reliable deployment, yet they have largely been studied in isolation.
The need for both generalization and adversarial robustness extends to emerging model architectures, including prototype-based models, which have recently gained attention in audio classification research \cite{zinemanas2021interpretable,ren2022prototype,liu2023interpretability,heinrich2025audioprotopnet}. 
In domains like bioacoustics, where model decisions often require expert verification, interpretability is a key requirement alongside classification accuracy. 
Architectures like \gls*{audioprotopnet} \cite{heinrich2025audioprotopnet} have demonstrated state-of-the-art performance in bird sound classification while providing the added benefit of global and local explanations through learned prototypes. 
However, the performance and interpretability of these models are critically dependent on the stability of their learned prototypes. 
Prototype sensitivity to distribution shifts or input perturbations can compromise classification and explanation reliability, making robust embeddings essential for deploying prototype-based models.

Given the inherent complexity and dynamic nature of audio signals, training strategies must evolve beyond conventional data augmentation. 
In this work, we investigate adversarial training as a principled augmentation strategy for multi-label bird sound classification under strong distribution shifts using BirdSet \cite{rauch2024birdset}.
Methodologically, we build on TRADES \cite{zhang2019theoretically} with adversarial weight perturbations (TRADES-AWP) \cite{wu2020adversarial} and adapt it to the multi-label setting by employing an asymmetric loss.
We compare two variants: \gls*{ato}, whose input perturbations maximize classification loss, and \gls*{ate}, whose input perturbations maximize embedding dissimilarity.
We evaluate these training strategies on two complementary architectures. 
ConvNeXt \cite{liu2022convnet} serves as a strong \gls*{cnn} baseline for bird sound classification \cite{rauch2024birdset}, and \gls*{audioprotopnet} \cite{heinrich2025audioprotopnet} is a prototype-based model that has demonstrated state-of-the-art performance in bird sound classification while providing prototype explanations.
Throughout the study, \gls*{ot} denotes ordinary training with the same preprocessing, data augmentations, loss, and optimizer configuration as the main experiments (Sections~\ref{subsec:data} and \ref{subsec:classification_models}), but with all adversarial training components disabled.
We evaluate how these training regimes affect classification performance and explanation stability when models transition from controlled training data to diverse real-world acoustics.

\subsection{Contributions}
This study systematically investigates adversarial training as an advanced augmentation strategy to improve generalization under distribution shifts and strengthen the adversarial robustness of \gls*{dl} models in multi-label audio classification.
Our investigation specifically focuses on bird sound classification, a domain often characterized by substantial distribution shifts between training data and real-world operational environments. 
The primary contributions of this work are as follows:
\begin{enumerate}
    \item[(C1)] We adapted the TRADES-AWP adversarial training framework \cite{wu2020adversarial} for multi-label classification by integrating an asymmetric loss and formulated two distinct variants thereof, which employ input perturbation attacks targeting either the model's output-space or its embedding-space.
    \item[(C2)]  We quantified the impact of output-space and embedding-space adversarial training on the clean-data generalization of both ConvNeXt, as a strong baseline, and \gls*{audioprotopnet}, as a state-of-the-art bird sound classification model with inherent interpretability.
    Our results demonstrate that output-space adversarial training enhances clean-data performance by up to 10.5\% relative in \glsentryshort{cmap} on average, compared to ordinary training. 
    Embedding-space adversarial training yielded more modest average improvements, with relative gains of up to 7.9\%.
    \item[(C3)] We comprehensively evaluated the adversarial robustness of models trained with ordinary and adversarial training methods against untargeted output-space and embedding-space attacks. 
    The results demonstrate that adversarial training enhances adversarial robustness, i.e., improves the retention of classification performance under norm-bounded input perturbations.
    For example, under untargeted embedding-space attacks with perturbation strength $\epsilon=0.01$, output-space adversarial training raised \gls*{audioprotopnet}'s \glsentryfull{prs} \cite{heinrich2024targeted} from 0.74 to 0.89 versus ordinary training.
    \item[(C4)] We specifically assessed the stability of prototypes learned by \gls*{audioprotopnet} under targeted embedding-space adversarial attacks.
    Our experiments show that output-space adversarial training substantially reinforces the stability of \gls*{audioprotopnet} prototypes. 
    For example, at a perturbation strength of $\epsilon=0.1$, the mean \glsentryfull{tars} \cite{heinrich2024targeted} improved from 0.27 with ordinary training to 0.65 with output-space adversarial training.
\end{enumerate}

\section{Related Work}
\label{sec:related_work}
Our work builds on output- and embedding-space adversarial training, adversarial training for distribution-shift generalization, and adversarial training for explanation robustness.

\textbf{Output-space adversarial training:}
Output-space adversarial training, which perturbs model inputs to maximize classification loss, is a widely studied defense strategy.
Seminal work by \citet{goodfellow2014explaining} introduced the \gls*{fgsm} to generate adversarial inputs in a single step, integrating them into the training process to minimize a mixed loss on clean and perturbed data.
Building on this, \citet{madry2018towards} formulated adversarial training as a minimax optimization problem, employing the multi-step \gls*{pgd} method to search for high-loss perturbations within a small norm-ball around the input.
The \gls*{trades} framework \cite{zhang2019theoretically} refines this by minimizing a loss that combines standard classification error with a term that penalizes divergence between classifications on clean and adversarial inputs.
Further extending these ideas, \citet{wu2020adversarial} proposed \gls*{awp}, which perturbs model weights in addition to inputs to flatten the loss landscape and improve how well adversarial robustness generalizes to unseen data.
They combined \gls*{awp} with TRADES in TRADES-AWP, which has been shown to further enhance adversarial robustness.
More recently, \citet{jin2023randomized} developed a defense that injects Gaussian noise into model weights, uses the resulting randomized model to generate adversarial inputs, and then minimizes a Taylor-expanded loss.
While these methods are well-established for multiclass classification, their adaptation to multi-label classification remains less explored \cite{rao2020thorough,xu2021towards,dong2024survey}.
In this work, we adapt TRADES-AWP for multi-label audio classification by replacing the cross-entropy divergence term with an asymmetric loss variant (Section~\ref{subsec:adv_training}).

\textbf{Embedding-space adversarial training:}
Embedding-space adversarial training offers an alternative to output-space adversarial training by targeting a model's embeddings.
While some methods directly perturb latent embeddings \cite{zhu2019freelb,xu2021towardsfeature,altinisik2022impact,qi2023robust}, this work focuses on crafting input perturbations that alter the similarity between embeddings of clean and perturbed inputs. 
This approach is chosen because input-level perturbations better reflect the practical challenges of real-world audio classification compared to direct embedding-level perturbations.
The primary motivation is to enforce desirable properties, such as local smoothness or semantic separation, directly in the latent embedding space, often in an unsupervised manner that avoids potential label leakage \cite{li2023adversarial,zhang2019defense}.
Much of the foundational work in embedding-space adversarial training originated in \gls*{ssl}, where input perturbations maximize a similarity-based objective between two different views of an instance before a contrastive loss restores their original embedding similarity \cite{kim2020adversarial,chen2020self,ho2020contrastive,jiang2020robust}.
Following a similar principle, \citet{zhang2024self} use a student-teacher framework where input and weight perturbations are used to disrupt embedding similarity with a frozen teacher, which the student model is then trained to restore.
These ideas have also been adapted for supervised learning frameworks. 
For example, feature-scattering adversarial training perturbs inputs to maximize the optimal-transport distance between the embedding distributions of clean and perturbed data \citep{zhang2019defense}. 
Similarly, \citet{li2023adversarial} extended supervised contrastive learning by using \gls*{pgd} to craft input perturbations that maximize the contrastive loss, effectively pushing embeddings of the same class further apart before the model is trained to pull them back together.
Despite this progress in computer vision, applications in the audio domain remain scarce. 
A notable exception is the work of \citet{li2024adversarial}, who improved the robustness of a text-to-speech system by using targeted embedding-space attacks to shift an utterance's speaker embedding toward that of another speaker, then fine-tuning the model to reverse this change.
However, a systematic comparison of output-space and embedding-space adversarial training in the audio domain remains an open research area.
We address this gap by providing the first such comparison for multi-label audio classification, evaluating both training variants on generalization, adversarial robustness, and prototype stability (Sections~\ref{subsec:performance_evaluation_clean}, \ref{subsec:robustness_untargeted_attacks}, and~\ref{subsec:robustness_prototypes}).

\textbf{Adversarial training for generalization under distribution shifts:}
Adversarial training can improve model generalization under distribution shifts, although its effect on clean-data performance is debated. 
Some studies report an inherent trade-off, where adversarial training improves adversarial robustness but harms performance on clean data, particularly in data-rich regimes \cite{tsipras2018robustness,javanmard2020precise,raghunathan2020understanding}. 
However, other work suggests that no such trade-off is fundamental, arguing that a locally Lipschitz classifier can achieve both high performance and adversarial robustness \cite{yang2020closer}. 
Empirical evidence shows that adversarial training can boost \gls*{ood} performance across various data distribution shifts, such as correlation and covariate shifts \cite{volpi2018generalizing,wang2022improving,xin2023connection}. 
Further studies support this, showing that adversarially trained models are more robust against naturally occurring perturbations \cite{gokhale2021attribute} and that adversarial robustness is theoretically linked to better \gls*{ood} generalization \cite{yi2021improved}. 
Additionally, models trained with \gls*{pgd} or \gls*{trades} have been found to retain their adversarial robustness on unseen domains \cite{alhamoud2022generalizability}. 
Despite these findings, the effects of output-space and embedding-space adversarial training on generalization and adversarial robustness in multi-label audio classification remain unexplored.
Our experiments directly investigate this question using the BirdSet benchmark, which provides a controlled setting with substantial, naturally occurring distribution shifts between focal training data and passive soundscape test data (Section~\ref{subsec:performance_evaluation_clean}).

\textbf{Adversarial training for explanation robustness:}
Adversarial methods are used not only to assess adversarial robustness but also to evaluate and improve the robustness of explanations from interpretable models.
The robustness of prototype-based explanations in particular has been scrutinized using various adversarial attacks.
Researchers have evaluated the robustness of the \gls*{ppnet} \cite{chen2019looks} against common image alterations \cite{rymarczyk2020protopshare}, untargeted output-space attacks \cite{yang2022multi, nakka2020towards}, and both untargeted and targeted embedding-space attacks designed to disrupt the model's learned prototype similarities \cite{sacha2024interpretability, hoffmann2021looks, huang2023evaluation}. 
Building on these evaluation methods, adversarial training has been explored as a technique to improve explanation robustness. 
Initial studies have shown that embeddings from adversarially trained models can align better with salient data characteristics and human perception \cite{tsipras2018robustness}. 
However, early attempts using \gls*{fgsm}-based adversarial training for prototype-based models revealed a trade-off, where enhanced robustness was accompanied by a significant drop in clean-data performance \cite{hoffmann2021looks, nakka2020towards}. 
The potential for modern adversarial training methods like \gls*{trades}, \gls*{awp}, or embedding-space adversarial training to mitigate this trade-off in prototype-based models remains an open research question.
We investigate exactly this by applying both TRADES-AWP-based output-space and embedding-space adversarial training to AudioProtoPNet and evaluating prototype stability under targeted attacks (Section~\ref{subsec:robustness_prototypes}).

\section{Methodology: Adversarial Attacks and Training for Multi-Label Audio Classification}
\label{sec:methodology}
This section formalizes the adversarial attack objectives and the adversarial training variants that we adapt and extend for multi-label audio classification.

\subsection{Adversarial attack strategies}
\label{subsec:attacks}
We assume a white-box setting, enabling direct gradient-based evaluation under controlled perturbation budgets.
We next define the three attacks used throughout the paper---untargeted output-space, untargeted embedding-space, and targeted embedding-space attacks---with additional background on threat models, attack objectives, and training in Appendix~\ref{sec:appendix_adversarial_background}.

\subsubsection{Untargeted output-space adversarial attacks}
\label{subsubsec:untargeted_output_space}
In audio classification, the raw waveform $\mathbf{s} \in \mathbb{R}^{T}$ is commonly transformed into a time-frequency representation such as a log-Mel spectrogram, because it compactly captures spectral structure and aligns well with perceptual frequency resolution.
We denote the resulting input spectrogram by $\mathbf{x}\in\mathbb{R}^{H\times W\times C}$, where $H$, $W$, and $C$ are its height, width, and number of channels.
A \gls*{dl} classification model $f_{\boldsymbol{\theta}}:\mathbb{R}^{H\times W\times C}\to[0,1]^K$, parameterized by $\boldsymbol{\theta}\in\mathbb{R}^d$, maps $\mathbf{x}$ to $K$ class confidence scores $\hat{\mathbf{y}}=f_{\boldsymbol{\theta}}(\mathbf{x})$ for the multi-label target $\mathbf{y}\in\{0,1\}^K$, with each score estimating the presence of one class.

An untargeted output-space attack seeks to degrade classification by finding a perturbation $\boldsymbol{\delta}$ that maximizes the supervised loss $\mathcal{L}:[0,1]^K\to\mathbb{R}^+$ used for ordinary training, thereby increasing the discrepancy between the model's classifications and the true labels.
For $\boldsymbol{\delta}$ in a permissible set $\mathcal{S}\subseteq\mathbb{R}^{H\times W\times C}$, typically an $\ell_\infty$-norm ball, this worst-case objective is
\begin{equation}
\max_{\boldsymbol{\delta} \in \mathcal{S}} \mathcal{L}\left(f_{\boldsymbol{\theta}}(\mathbf{x} + \boldsymbol{\delta}), \mathbf{y}\right).
\label{eq:equation_untargeted_output}
\end{equation}
We generate such perturbations with \gls*{pgd} \cite{madry2018towards}, an iterative gradient-based method that is effective because it uses white-box access to the loss gradient with respect to the input \cite{carlini2019evaluating} while projecting each update back into the permissible set.
\gls*{pgd} supports both targeted and untargeted objectives and, here, maximizes the discrepancy between the model's classifications and the true labels.
The common projected-gradient update and its objective-specific instantiations are provided in Appendix~\ref{sec:appendix_pgd_optimization}.

\subsubsection{Embedding-space adversarial attacks}
\label{subsubsec:embedding_space_attacks}
\gls*{dl} architectures for audio classification typically employ an embedding extractor $f_{\boldsymbol{\theta}_b}^b: \mathbb{R}^{H \times W \times C} \to \mathbb{R}^{H_z \times W_z \times D}$ and a subsequent classifier $f_{\boldsymbol{\theta}_c}^c: \mathbb{R}^{H_z \times W_z \times D} \to [0,1]^K$. 
The embedding extractor maps an input spectrogram $\mathbf{x}$ to a latent map $\mathbf{z} \in \mathbb{R}^{H_z \times W_z \times D}$, whose height, width, and channel count are $H_z$, $W_z$, and $D$, and which provides an internal representation for downstream tasks such as classification or retrieval \cite{rauch2025can, dumoulin2025search}.
The overall model can be expressed as $f_{\boldsymbol{\theta}} = f_{\boldsymbol{\theta}_c}^c \circ f_{\boldsymbol{\theta}_b}^b$, where $\circ$ denotes function composition.
Directly manipulating these latent maps through adversarial attacks reveals vulnerabilities distinct from conventional output-space attacks: although $\|\boldsymbol{\delta}\|_\infty$ is small, the approximately linear behavior of deep networks in high-dimensional input spaces \cite{goodfellow2014explaining} allows small perturbations to accumulate large effects across dimensions, potentially amplifying norm-bounded perturbations into substantial changes in intermediate representations \cite{szegedy2013intriguing,heinrich2024targeted}.
This susceptibility not only hurts classification performance but also poses a significant problem for applications that depend on embedding integrity, such as retrieval systems in bioacoustics \cite{dumoulin2025search}.
In the following, we delineate both untargeted and targeted embedding-space attacks formulated within a white-box setting employing gradient-based optimization.

\paragraph{Untargeted embedding-space adversarial attacks.}
Untargeted embedding-space attacks seek an input perturbation $\boldsymbol{\delta}\in\mathcal{S}$ that maximizes the dissimilarity between the embedding maps generated from the clean and the perturbed inputs.
Let $\mathbf{z} = f_{\boldsymbol{\theta}_b}^b(\mathbf{x})$ be the original embedding map and $\mathbf{z}_{adv} = f_{\boldsymbol{\theta}_b}^b(\mathbf{x}+\boldsymbol{\delta})$ be the embedding map of the perturbed input. 
The attack objective is to maximize a dissimilarity function $\mathcal{D}(\mathbf{z}_{adv}, \mathbf{z})$. This objective is formally expressed as
\begin{equation}
    \max_{\boldsymbol{\delta} \in \mathcal{S}} \mathcal{D} \left( f_{\boldsymbol{\theta}_b}^b (\mathbf{x} + \boldsymbol{\delta}), \mathbf{z} \right).
    \label{eq:equation_untargeted_embedding}
\end{equation}

\begin{figure}[!ht]
    \centering
    \begin{subfigure}[b]{0.25\textwidth}
        \centering
        \includegraphics[width=0.95\textwidth]{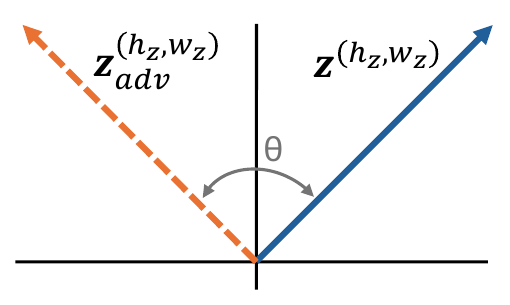}
        \caption{}
    \end{subfigure}
    \hspace{1cm}
    \begin{subfigure}[b]{0.25\textwidth}
        \centering
        \includegraphics[width=0.95\textwidth]{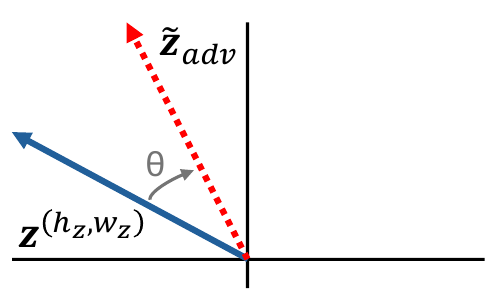}
        \caption{}
    \end{subfigure}
    \caption{Embedding-space attacks perturb model inputs to manipulate the resulting embeddings, either by disrupting their structure or by aligning them with a chosen target. (a) Untargeted attacks maximize the average spatial cosine distance $\mathcal{D}_{avg}$ between original embedding vectors $\mathbf{z}^{(h_z, w_z)}$ and adversarial embedding vectors $\mathbf{z}_{adv}^{(h_z, w_z)}$ aiming for orthogonality. (b) Targeted attacks minimize the minimum spatial cosine distance $\mathcal{D}_{min}$ between embedding vectors $\mathbf{z}_{adv}^{(h_z, w_z)}$ generated from the perturbed input and a target embedding vector $\tilde{\mathbf{z}}_{adv}$ aiming for alignment.
    }   \label{fig:embedding_space_attacks}
\end{figure}

Possible choices for the dissimilarity function $\mathcal{D}$ include cosine distance and squared Euclidean distance.
Prior work has used cosine similarity to compare spatial embeddings for embedding-space robustness analysis \cite{sacha2024interpretability}.
In our formulation, we instantiate $\mathcal{D}$ as a spatial extension of cosine distance.
Specifically, for untargeted embedding-space attacks, we define the average spatial cosine distance $\mathcal{D}_{avg}$ between the clean embedding map $\mathbf{z}$ and the adversarial embedding map $\mathbf{z}_{adv}$, as illustrated in Figure~\ref{fig:embedding_space_attacks}.
Cosine distance is well suited for high-dimensional embeddings because it measures the angle between vectors, making it robust to magnitude variations \cite{luo2018cosine,you2025semantics}.
We define $\mathcal{D}_{avg}$ as:
\begin{equation}
  \mathcal{D}_{avg}\left( \mathbf{z}, \mathbf{z}_{adv} \right)
  = \frac{1}{H_z W_z} \sum_{h_z=1}^{H_z} \sum_{w_z=1}^{W_z} 
    \left( 1 - \max\!\left(0,\;
      \frac{\mathbf{z}^{(h_z, w_z)} \cdot \mathbf{z}_{adv}^{(h_z, w_z)}}{\left\|\mathbf{z}^{(h_z, w_z)}\right\|_2 \cdot \left\|\mathbf{z}_{adv}^{(h_z, w_z)}\right\|_2}
    \right) \right).
  \label{eq:avg_spatial_cosine_distance}
\end{equation}
At every spatial position $(h_z,w_z)$, where $1\leq h_z\leq H_z$ and $1\leq w_z\leq W_z$, we compute the cosine similarity between the corresponding local vectors $\mathbf{z}^{(h_z,w_z)}$ and $\mathbf{z}_{adv}^{(h_z,w_z)}$.
The \gls*{relu} \cite{nair2010rectified} truncates negative similarities to zero, so substantially different or opposing vectors are treated as maximally dissimilar at that position.
Subtracting the rectified similarity from one gives the local dissimilarity, whose arithmetic mean over all $H_zW_z$ positions, $\mathcal{D}_{avg}$, quantifies the perturbation's average effect across the complete embedding map.
We maximize $\mathcal{D}_{avg}$ using the \gls*{pgd} procedure summarized in Appendix~\ref{sec:appendix_pgd_optimization}.

\paragraph{Targeted embedding-space adversarial attacks.}
Targeted embedding-space adversarial attacks seek a perturbation $\boldsymbol{\delta}\in\mathcal{S}$ that makes the resulting embedding map similar to a predefined target embedding vector $\tilde{\mathbf{z}}_{adv}\in\mathbb{R}^{D}$.
Formally, the local vectors of $\mathbf{z}_{adv}=f_{\boldsymbol{\theta}_b}^b(\mathbf{x}+\boldsymbol{\delta})$ are driven towards $\tilde{\mathbf{z}}_{adv}$ by minimizing their dissimilarity:
\begin{equation}
    \min_{\boldsymbol{\delta} \in \mathcal{S}} \mathcal{D} \left( f_{\boldsymbol{\theta}_b}^b (\mathbf{x} + \boldsymbol{\delta}), \tilde{\mathbf{z}}_{adv} \right).
\end{equation}
Thus, although the perturbed input $\mathbf{x}_{adv}$ may remain similar to the original input, its latent embedding is encouraged to align with $\tilde{\mathbf{z}}_{adv}$.
For these attacks, we use the minimum spatial cosine distance $\mathcal{D}_{min}$, illustrated in Figure~\ref{fig:embedding_space_attacks}.

$\mathcal{D}_{min}$ identifies the spatial position at which a local adversarial embedding vector is least dissimilar to the target.
It is defined as the minimum of the local cosine dissimilarities:
\begin{equation}
  \mathcal{D}_{min} \left( \mathbf{z}_{adv}, \tilde{\mathbf{z}}_{adv} \right)
  = \min_{h_z, w_z} \left\{ 1 - \max\!\left(0,\;
    \frac{\mathbf{z}_{adv}^{(h_z, w_z)} \cdot \tilde{\mathbf{z}}_{adv}}{\left\|\mathbf{z}_{adv}^{(h_z, w_z)}\right\|_2 \cdot \left\|\tilde{\mathbf{z}}_{adv}\right\|_2}
  \right) \right\}.
  \label{eq:min_spatial_cosine_distance}
\end{equation}
At each spatial position, the rectified cosine similarity between $\mathbf{z}_{adv}^{(h_z,w_z)}$ and $\tilde{\mathbf{z}}_{adv}$ is converted to a dissimilarity in the same manner as for $\mathcal{D}_{avg}$.
Taking the minimum over all positions means that a low $\mathcal{D}_{min}$ indicates that at least one local vector closely matches the target.
We minimize $\mathcal{D}_{min}$ using the \gls*{pgd} procedure summarized in Appendix~\ref{sec:appendix_pgd_optimization}, thereby guiding at least one spatial region of the adversarial embedding map towards $\tilde{\mathbf{z}}_{adv}$.

The target $\tilde{\mathbf{z}}_{adv}$ depends on the application and attacker's intent. For \gls*{audioprotopnet} \cite{heinrich2025audioprotopnet}, it can be a learned class prototype $\mathbf{p}_k\in\mathbb{R}^{D}$ that the attack seeks to activate strongly, or the pooled embedding map $f_{\boldsymbol{\theta}_b}^b(\mathbf{x}_n)$ of an audio instance $\mathbf{x}_n$ whose characteristics the attacker seeks to impose.
Such attacks assess prototype stability and thus the reliability of \gls*{audioprotopnet}'s explanations.

\subsection{Adversarial training}
\label{subsec:adv_training} 
We use the TRADES-AWP framework \cite{wu2020adversarial}, which extends standard adversarial training \cite{madry2018towards} by combining \gls*{trades} \cite{zhang2019theoretically} and \gls*{awp} \cite{wu2020adversarial} in a unified objective.
\gls*{trades} improves the balance between clean-data performance and adversarial robustness by adding a regularization term that penalizes inconsistencies between model outputs for clean and perturbed inputs \cite{zhang2019theoretically}.
\gls*{awp} further enhances robustness by perturbing model weights directly, which promotes a flatter loss landscape and reduces the gap between robustness on training data and on unseen test data \cite{wu2020adversarial}.

\paragraph{Multi-label adaptation via asymmetric loss.}
In this work we adapt TRADES-AWP to multi-label bird sound classification by (i) replacing the multiclass loss with the asymmetric multi-label loss $\mathcal{L}_{asym}$ \cite{ridnik2021asymmetric}, (ii) introducing two instantiations of the inner input-perturbation step, namely output-space attacks (AT-O, Eq.~\ref{eq:equation_untargeted_output}) and embedding-space attacks (AT-E, Eq.~\ref{eq:equation_untargeted_embedding}), and (iii) approximating the inner maximization during training with single-step FGSM for efficiency while applying weight randomization during perturbation generation following \citet{jin2023randomized}.
The combined objective function for our adversarial training approach is:
\begin{equation}
\min_{\boldsymbol{\theta}} \;\max_{\mathbf{v}:\,||\mathbf{v}_{j}|| \le \gamma ||\boldsymbol{\theta}_{j}||} \;\mathbb{E}_{(\mathbf{x},\mathbf{y})\sim\mathcal{F}}\left[    \mathcal{L}\left(f_{\boldsymbol{\theta}+\mathbf{v}}(\mathbf{x}),\mathbf{y}\right)    + \frac{1}{\lambda}\,    \mathcal{L}\left(f_{\boldsymbol{\theta}+\mathbf{v}}(\mathbf{x}+\boldsymbol{\delta}), f_{\boldsymbol{\theta}+\mathbf{v}}(\mathbf{x})\right)\right].
\label{eq:unified_adversarial_training}
\end{equation}
Here $(\mathbf{x},\mathbf{y})\sim\mathcal{F}$ denotes the training data, $\mathbf{v}$ is the adversarial weight perturbation, $\boldsymbol{\theta}_{j}$ and $\mathbf{v}_{j}$ denote respectively the weight tensor and its perturbation in layer $j$, and $\gamma>0$ defines the relative layer-wise perturbation budget.
The supervised loss $\mathcal{L}$ is asymmetric to address label imbalance: it down-weights easy negatives, i.e., absent labels predicted with low confidence, while emphasizing positives and hard negatives, i.e., absent labels assigned relatively high confidence and thus large multi-label training loss \cite{ridnik2021asymmetric}.
In the objective function, $\mathcal{L}$ is applied first to measure the supervised classification error $\mathcal{L}(f_{\boldsymbol{\theta}+\mathbf{v}}(\mathbf{x}),\mathbf{y})$. 
It is then used in the robustness regularization term $\mathcal{L}(f_{\boldsymbol{\theta}+\mathbf{v}}(\mathbf{x}+\boldsymbol{\delta}), f_{\boldsymbol{\theta}+\mathbf{v}}(\mathbf{x}))$ to quantify the dissimilarity between the model's output distributions for clean and perturbed inputs, where the output distribution of the clean input serves as the target.
The perturbation $\boldsymbol{\delta}$ is confined to a set of allowable perturbations $\mathcal{S}=\{\boldsymbol{\delta}\in\mathbb{R}^{H \times W \times C}\!:\|\boldsymbol{\delta}\|_{\infty}\le\epsilon\}$. 
The hyperparameter $\lambda > 0$ controls the trade-off between classification loss and the robustness regularization term.
The generation of the perturbation $\boldsymbol{\delta}$ follows the untargeted output-space attack objective in Equation~\ref{eq:equation_untargeted_output} for output-space adversarial training and the untargeted embedding-space attack objective in Equation~\ref{eq:equation_untargeted_embedding} for embedding-space adversarial training.

\paragraph{Efficient inner maximization.}
The optimal $\boldsymbol{\delta}$ could be approximated with the generic \gls*{pgd} update in Appendix~\ref{sec:appendix_pgd_optimization} and the corresponding objective \cite{madry2018towards}.
Although multi-step \gls*{pgd} typically identifies more potent adversarial inputs, approximating the optimum at each training step is computationally intensive because its iterative nature significantly increases training time.
To maintain computational efficiency in our experiments, we therefore approximate the perturbation generation using the untargeted \gls*{fgsm} attack \cite{goodfellow2014explaining} for both output-space and embedding-space adversarial training. 
\gls*{fgsm} is a single-step variant of \gls*{pgd} \cite{golgooni2021zerograd,tsiligkaridis2022understanding}, which offers a computationally efficient yet effective method for adversarial training \cite{goodfellow2014explaining,wong2020fast,andriushchenko2020understanding}.
To further bolster robustness, we also randomize model weights during the generation of input perturbations, following the methodology of \citet{jin2023randomized}.
The outer minimization problem then adjusts the model parameters to enhance robustness against these specific perturbations, a step that is resolved using a standard training procedure.
This comprehensive training strategy is designed to yield models that are not only accurate on clean data but also maintain their performance under various adversarial conditions.

\section{Experimental Setup for Bird Sound Classification}
\label{sec:experiments}

\subsection{Datasets and preprocessing}
\label{subsec:data}
We use BirdSet \cite{rauch2024birdset}, a standardized evaluation framework for bird sound classification in bioacoustics that mirrors the challenges of \gls*{pam} by training models on curated focal recordings and evaluating them on continuous soundscape recordings captured by stationary sensors deployed in the field.
Focal recordings typically feature isolated, high-quality vocalizations, whereas soundscape recordings are characterized by heterogeneous recording devices, diverse background noise, and overlapping species vocalizations \cite{rauch2024birdset}.
This realistic shift and class imbalance across seven soundscape test datasets (Appendix~\ref{sec:appendix_preprocessing}) make BirdSet well suited to studying how adversarial training affects generalization.

We adopt its dedicated training task, in which a Xeno-Canto \cite{vellinga2015xeno} training subset containing only the respective target classes is curated for each soundscape test dataset.
We choose this task over large-scale training across all species because each model is trained exclusively on the species on which it is evaluated, allowing performance differences to be attributed to the domain gap between focal recordings and soundscape data rather than to unseen classes.
To select the adversarial perturbation strength $\epsilon$, we conduct an ablation study using the \gls*{pow} \cite{powdermill_chronister_2021_4656848} training and test datasets as the dedicated training and validation sets (Section~\ref{subsec:impact_of_perturbation_strength}).

Audio data are transformed into log-Mel spectrograms following the preprocessing protocol of \citet{heinrich2025audioprotopnet} and standardized via z-score normalization.
Preprocessing and augmentation follow \citet{rauch2024birdset} and \citet{heinrich2025audioprotopnet}; Table~\ref{tab:preprocessing_augmentation_params} in Appendix~\ref{sec:appendix_preprocessing} lists all parameters.

\subsection{Classification models and training}
\label{subsec:classification_models}

\paragraph{Model architectures.}
We evaluate two complementary \gls*{cnn}-based architectures that share a ConvNeXt-B \cite{liu2022convnet} backbone but differ in their classification heads.
ConvNeXt-B serves as a strong baseline that has shown competitive performance in bird sound classification \cite{rauch2024birdset}.
It maps spectrograms through its convolutional backbone to latent embedding maps (as formalized in Section~\ref{subsubsec:embedding_space_attacks}), which are then processed by a linear classification head.
\gls*{audioprotopnet} \cite{heinrich2025audioprotopnet} extends the same backbone with a prototype-based classification head instead of a linear layer.
Classification is based on the similarity between input embedding maps and these prototypes, thereby providing interpretable explanations for individual predictions.
The complete architecture specifications are provided in Table~\ref{tab:training_hyperparams} in Appendix~\ref{sec:appendix_training_config}.

\paragraph{Ordinary training.}
Each model is fine-tuned on the dedicated BirdSet training subsets described in Section~\ref{subsec:data} and Appendix~\ref{sec:appendix_preprocessing}.
Both architectures use the asymmetric loss $\mathcal{L}_{asym}$ \cite{ridnik2021asymmetric} to address class imbalance.
\gls*{ot} denotes the standard non-adversarial training recipe, which uses the preprocessing, augmentations, and optimization configuration described in Section~\ref{subsec:data}, Appendix~\ref{sec:appendix_preprocessing} and Appendix~\ref{sec:appendix_training_config}, without adversarial input perturbations or TRADES-AWP regularization.

\paragraph{Adversarial training.}
In addition to \gls*{ot}, we train both architectures using the TRADES-AWP framework described in Section~\ref{subsec:adv_training}.
\gls*{ato} employs untargeted output-space \gls*{fgsm} attacks, whereas \gls*{ate} employs untargeted embedding-space \gls*{fgsm} attacks.
Weight randomization \cite{jin2023randomized} is applied during adversarial input generation.
For \gls*{ate}, the clean reference embedding is computed with the unperturbed model before this randomization and held fixed during FGSM generation.
We selected the perturbation strength $\epsilon$ for \gls*{fgsm} attacks via an ablation study on the \gls*{pow} dataset (Section~\ref{subsec:impact_of_perturbation_strength}), evaluating $\epsilon \in \{0.001, 0.01, 0.05, 0.1, 0.2\}$.
The complete optimization, adversarial-training, and model-selection specifications are provided in Table~\ref{tab:training_hyperparams}.

\subsection{Evaluation metrics}
\label{subsec:evaluation_metrics}

\paragraph{Performance.}
Following \citet{rauch2025can}, we primarily assessed performance using \gls*{cmap}, which averages \gls*{ap} across all $K$ classes and is therefore equitable across class frequencies and well suited to multi-label audio classification \cite{kahl2023overview}.
The formal definition and supporting notation are provided in Appendix~\ref{subsec:appendix_cmap_definition}.
Additionally, we report \gls*{auroc} and Top-1 Accuracy in Appendix~\ref{subsec:appendix_additional_metrics}.

\paragraph{Adversarial robustness.}
To evaluate empirical robustness under bounded input perturbations, we used \gls*{pgd} attacks comprising untargeted output-space attacks as well as both untargeted and targeted embedding-space attacks.
Perturbations were confined to the $\ell_\infty$-norm ball with maximum magnitudes $\epsilon \in \{0.001, 0.01, 0.05, 0.1, 0.2\}$, and each attack was run for $T = 10$ iterations with step size $\alpha = \frac{2\epsilon}{T}$.
Each evaluation PGD attack used a single random input initialization within the corresponding $\ell_\infty$ ball.
For untargeted output-space and embedding-space attacks, we quantified robustness with the \gls*{prs} \cite{heinrich2024targeted}; for targeted embedding-space attacks, we additionally used the \gls*{drs} and \gls*{tars} \cite{heinrich2024targeted}.
The formal definitions and supporting notation are provided in Appendix~\ref{subsec:appendix_robustness_metrics_definitions}.

\section{Results}
\label{sec:results} 
We evaluated the effects of adversarial training along three axes: (i) clean-data performance across varying perturbation budgets and geographically diverse test datasets (Sections~\ref{subsec:impact_of_perturbation_strength} and~\ref{subsec:performance_evaluation_clean}), (ii) adversarial robustness under untargeted attacks (Section~\ref{subsec:robustness_untargeted_attacks}), and (iii) prototype stability under targeted embedding-space attacks (Section~\ref{subsec:robustness_prototypes}).
Recall that \gls*{ot} denotes ordinary training without adversarial components, while \gls*{at} encompasses both \gls*{ato} (output-space) and \gls*{ate} (embedding-space) variants.

\subsection{Impact of perturbation strength on clean-data performance}
\label{subsec:impact_of_perturbation_strength}
We investigated how the adversarial training perturbation strength $\epsilon$ affects performance on clean, unperturbed audio using the POW validation dataset.
For both ConvNeXt and \gls*{audioprotopnet}, we compared the \gls*{cmap} after \gls*{ot} and \gls*{at} with five perturbation budgets $\epsilon \in \{0.001, 0.01, 0.05, 0.1, 0.2\}$.
Table~\ref{tab:clean_performance_POW} in Appendix~\ref{subsec:appendix_perturbation_ablation} presents the results.
Adversarial training substantially enhanced clean-data performance for both model architectures compared to ordinary training.
Without adversarial training, \gls*{audioprotopnet} achieved the higher baseline \gls*{cmap} of 0.49, compared with 0.47 for ConvNeXt.
For ConvNeXt, \gls*{ato} yielded the largest absolute gain: the medium perturbation budget $\epsilon=0.1$ increased \gls*{cmap} from 0.47 with \gls*{ot} to 0.56, a relative improvement of 19.1\%.
In contrast, \gls*{ate} produced a \gls*{cmap} of approximately 0.54 across the tested perturbation budgets, a notable relative improvement of 14.9\% over \gls*{ot}.
\gls*{audioprotopnet} also benefited from both strategies: \gls*{ate} increased its \gls*{cmap} from 0.49 to 0.57, a relative increase of 16.3\%, while \gls*{ato} produced an even greater improvement to 0.59 for $\epsilon \in \{0.05, 0.1, 0.2\}$, a relative increase of 20.4\% over ordinary training.
These findings on the POW dataset demonstrated that appropriately calibrated adversarial training can improve clean-data performance, indicating that adversarial training does not necessarily induce a trade-off between clean-data performance and adversarial robustness.
We therefore selected $\epsilon=0.1$ for subsequent \gls*{ato} and \gls*{ate} experiments because it consistently improved performance across architectures and adversarial training variants.

\subsection{Generalization to diverse soundscape test datasets} 
\label{subsec:performance_evaluation_clean}
\begin{figure}[!ht]
  \centering
  \includegraphics[scale=0.55]{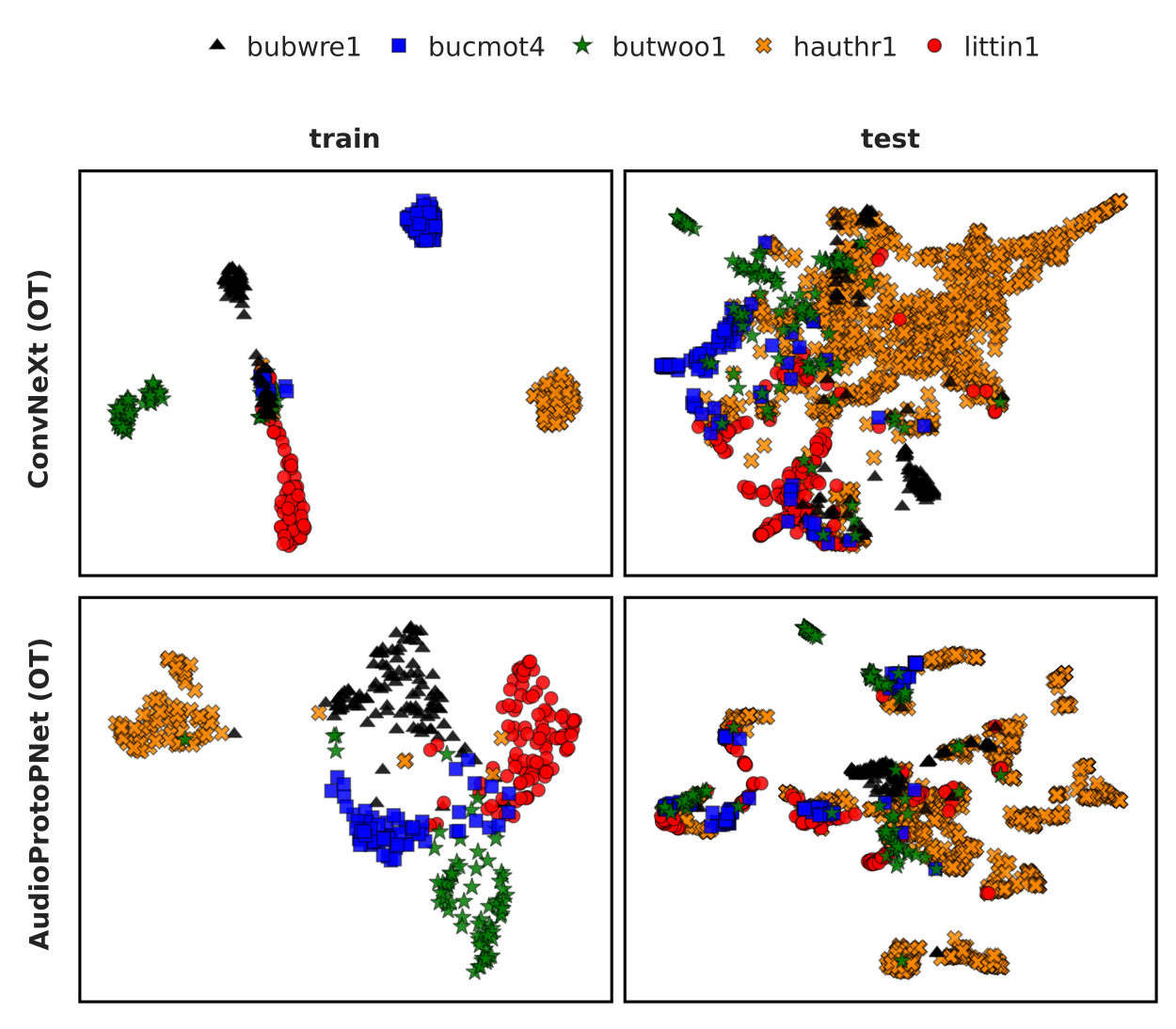}
  \caption{UMAP of global average pooled backbone embeddings (the spatially averaged ConvNeXt-B feature-extractor output before the classification head) from ordinarily trained ConvNeXt (top) and AudioProtoPNet (bottom) for single-label instances of the five most common PER bird species. Training embeddings (left) form distinct class clusters, whereas test embeddings (right) overlap substantially, confirming the train--test distribution shift.}
  \label{fig:embeddings_train_test}
\end{figure}

Having selected $\epsilon=0.1$, we next assessed generalization performance across seven geographically and acoustically diverse soundscape test datasets from BirdSet \cite{rauch2024birdset}. 
Table~\ref{tab:clean_performance} presents the \gls*{cmap} scores for both ConvNeXt and \gls*{audioprotopnet} under ordinary training and both adversarial training variants with $\epsilon=0.1$. 
\begin{table}[!ht]
  \caption{Clean-data \gls*{cmap} for ConvNeXt and \gls*{audioprotopnet} under \gls*{ot}, \gls*{ato}, and \gls*{ate} ($\epsilon=0.1$) across seven test datasets and their mean (Score). Parentheses show absolute gains over the corresponding \gls*{ot} baseline; bold and underlined values mark the best and second-best result per dataset.}
  \label{tab:clean_performance}
  \centering
  \resizebox{\textwidth}{!}{%
    \begin{tabular}{lccccccc|c}
    \toprule
    & \multicolumn{7}{c|}{\textbf{Test datasets}} & \\
    \cmidrule(lr){2-8}\cmidrule(lr){9-9}
    \textbf{Model}
      & \textbf{PER} & \textbf{NES} & \textbf{UHH}
      & \textbf{HSN} & \textbf{NBP} & \textbf{SSW} & \textbf{SNE}
      & \textbf{Score} \\
    \midrule
    ConvNeXt (OT)
      & 0.25    & 0.34    & 0.24    & 0.46    & \underline{0.68} & 0.37    & 0.31    & 0.38 \\
    \midrule
    ConvNeXt (AT-E)
      & 0.29\,{\scriptsize(+.04)}    & 0.37\,{\scriptsize(+.03)}    & 0.23\,{\scriptsize(-.01)}    & 0.43\,{\scriptsize(-.03)}    & \underline{0.68}\,{\scriptsize(.00)} & \underline{0.41}\,{\scriptsize(+.04)} & 0.32\,{\scriptsize(+.01)}    & 0.39\,{\scriptsize(+.01)} \\
    \midrule
    ConvNeXt (AT-O)
      & \underline{0.31}\,{\scriptsize(+.06)} & \underline{0.39}\,{\scriptsize(+.05)} & 0.23\,{\scriptsize(-.01)}    & 0.47\,{\scriptsize(+.01)}    & \textbf{0.69}\,{\scriptsize(+.01)}    & \textbf{0.44}\,{\scriptsize(+.07)}    & \underline{0.34}\,{\scriptsize(+.03)} & \underline{0.41}\,{\scriptsize(+.03)} \\
    \midrule
    AudioProtoPNet (OT)
      & 0.25    & 0.36    & \underline{0.25} & 0.46    & \underline{0.68} & 0.37    & 0.30    & 0.38 \\
    \midrule
    AudioProtoPNet (AT-E)
      & 0.30\,{\scriptsize(+.05)}    & 0.38\,{\scriptsize(+.02)}    & \textbf{0.26}\,{\scriptsize(+.01)}    & \underline{0.48}\,{\scriptsize(+.02)} & \textbf{0.69}\,{\scriptsize(+.01)}    & \underline{0.41}\,{\scriptsize(+.04)} & 0.32\,{\scriptsize(+.02)}    & \underline{0.41}\,{\scriptsize(+.03)} \\
    \midrule
    AudioProtoPNet (AT-O)
      & \textbf{0.32}\,{\scriptsize(+.07)}    & \textbf{0.40}\,{\scriptsize(+.04)}    & \underline{0.25}\,{\scriptsize(.00)} & \textbf{0.50}\,{\scriptsize(+.04)}    & \underline{0.68}\,{\scriptsize(.00)} & \textbf{0.44}\,{\scriptsize(+.07)}    & \textbf{0.35}\,{\scriptsize(+.05)}    & \textbf{0.42}\,{\scriptsize(+.04)} \\
    \bottomrule
  \end{tabular}
  }
\end{table}
Adversarial training generally improved or maintained performance relative to ordinary training, except for a marginal one-percentage-point \gls*{cmap} decrease for ConvNeXt on UHH.
Across the seven test datasets, \gls*{ato} increased the mean \gls*{cmap} from 0.38 to 0.41 for ConvNeXt, a relative gain of 7.9\%, and from 0.38 to 0.42 for \gls*{audioprotopnet}, a relative gain of 10.5\%.
Both architectures benefited slightly more from \gls*{ato} than \gls*{ate}, which increased the mean \gls*{cmap} from 0.38 to 0.39 for ConvNeXt and from 0.38 to 0.41 for \gls*{audioprotopnet}, corresponding to improvements of 2.6\% and 7.9\%, respectively.
These averages show a consistent advantage for adversarial training.
We observed the most substantial gains on the challenging PER dataset, which is characterized by significant background noise and overlapping vocalizations \cite{rauch2024birdset}.
To illustrate the distribution shift in the PER dataset, we visualized the global average pooled embeddings from the ordinarily trained ConvNeXt and AudioProtoPNet models, as shown in Figure~\ref{fig:embeddings_train_test}. 
Using default UMAP parameters \cite{mcinnes2018umap}, we projected the high-dimensional embeddings of single-label instances from the five most frequent PER bird species into two dimensions.
Training embeddings formed mostly distinct class clusters, whereas test embeddings had a different structure with considerable class overlap.
On PER, \gls*{ato} increased \gls*{audioprotopnet}'s \gls*{cmap} from 0.25 to 0.32 (28\%) and similarly increased ConvNeXt's from 0.25 to 0.31 (24\%).
Conversely, gains saturated on NBP, where \gls*{ot} already achieved a high \gls*{cmap} of 0.68, indicating diminishing returns when baseline performance was strong.
Nevertheless, adversarial training clearly improved average clean-data performance across the diverse test datasets.

To examine whether this dataset-dependent variation was associated with the magnitude of the train--test distribution shift, we estimated the shift for each of the seven BirdSet subsets using the mean per-class Fr\'echet distance in the Perch~v2 \cite{van2025perch} embedding space (see Appendix~\ref{sec:appendix_distribution_shift} for details).
For each subset and adversarial-training variant, we computed the change in \gls*{cmap} relative to the corresponding ordinary-training baseline separately for ConvNeXt and AudioProtoPNet and then averaged the two changes with equal weight (Table~\ref{tab:clean_performance}).
The Spearman rank correlation between these model-averaged gains and the corresponding shift estimates was $\rho_s = 0.75$ for \gls*{ate} and $\rho_s = 0.43$ for \gls*{ato}.
Although \gls*{ato} yielded larger gains overall, the relationship between the gains and the estimated distribution shift was stronger for \gls*{ate}: subsets with larger shifts generally exhibited larger \gls*{ate} gains.
Overall, larger estimated train--test distribution shifts tended to coincide with larger model-averaged gains from adversarial training, suggesting that adversarial training can be particularly beneficial when the train--test distribution shift is pronounced.

\maintextfloatbreak
\subsection{Adversarial robustness to untargeted attacks} 
\label{subsec:robustness_untargeted_attacks}
\begin{figure}[!ht]
    \centering
    \begin{subfigure}[b]{\textwidth}
        \centering
        \includegraphics[trim={0cm 0cm 0cm 0cm},clip,width=0.93\textwidth]{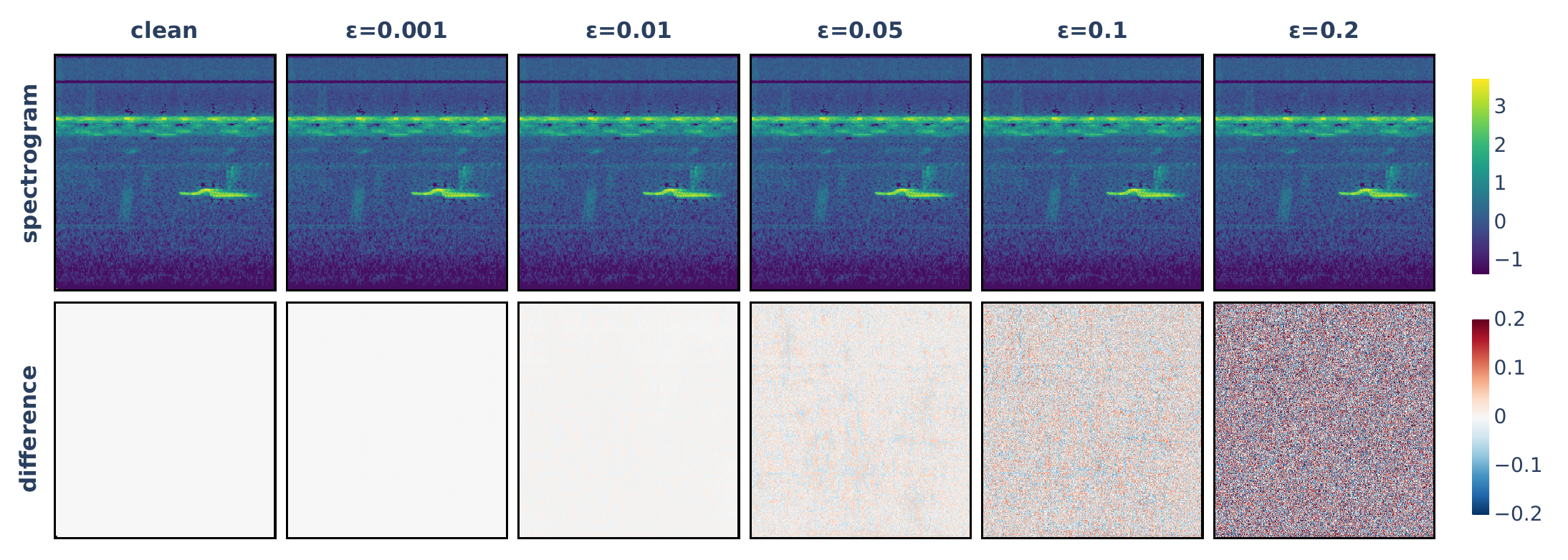}
        \caption{} 
        \label{fig:perturbations_untargeted_os_attacks_train}
    \end{subfigure}
    \vspace{1em}
    \begin{subfigure}[b]{\textwidth}
        \centering
        \includegraphics[trim={0cm 0cm 0cm 0cm},clip,width=0.93\textwidth]{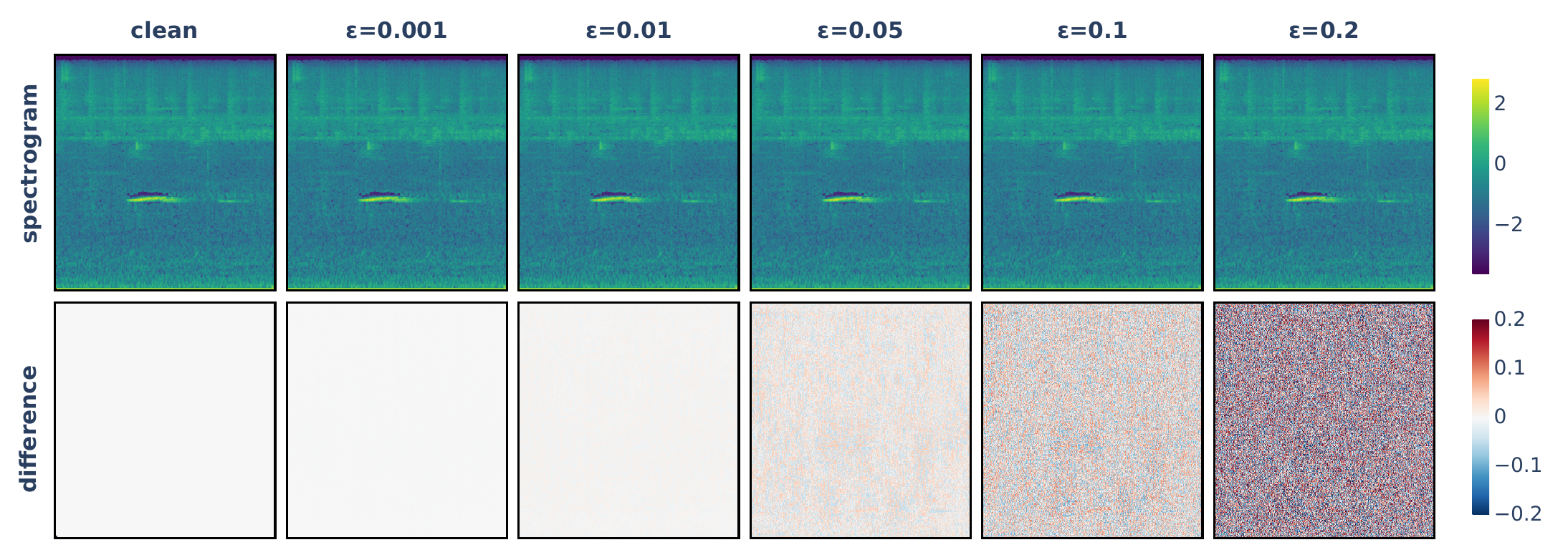}
        \caption{} 
        \label{fig:perturbations_untargeted_os_attacks_test}
    \end{subfigure}
    \caption{Clean and adversarially perturbed spectrograms with difference maps (perturbed minus clean) for untargeted output-space attacks of varying strength $\epsilon$ on an ordinarily trained ConvNeXt. Examples show a training instance (a) and test instance (b) from PER for Little Tinamou (littin1). Despite affecting model performance, perturbations were visually subtle, especially for the test instance.}
    \label{fig:perturbations_untargeted_os_attacks}
\end{figure}

\begin{figure}[!ht]
    \centering
    \includegraphics[trim={0cm 0cm 0cm 0cm},clip,width=\textwidth]{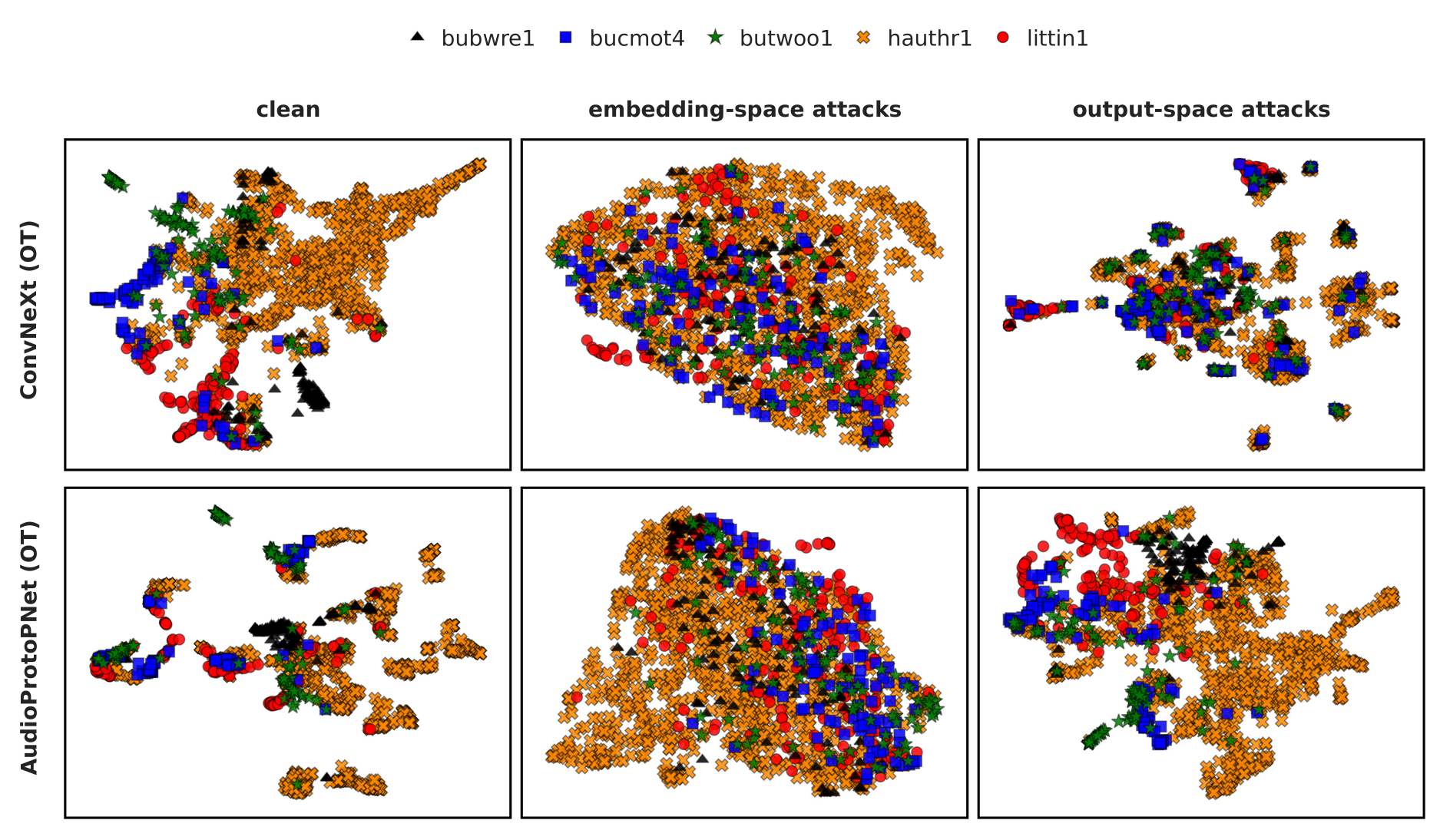}
    \caption{UMAP embeddings from ordinarily trained ConvNeXt (top) and \gls*{audioprotopnet} (bottom) for single-label instances of the five most common PER bird species. Clean inputs (left) are compared with inputs perturbed by untargeted embedding-space (middle) and output-space (right) \gls*{pgd} attacks at $\epsilon=0.05$. Embedding-space attacks destroy embedding structure, whereas output-space attacks degrade performance without fully collapsing the clusters.}
    \label{fig:embeddings_untargeted_attacks}
\end{figure}

We first assessed robustness against untargeted embedding-space \gls*{pgd} attacks, with Table~\ref{tab:robustness_untargeted_embedding_space} reporting \gls*{prs} across five perturbation strengths $\epsilon$.
At $\epsilon=0.001$, all models remained virtually unaffected, with \gls*{prs} near or at 1.0, suggesting that inherent data variation and standard augmentation provided sufficient robustness to such minor attacks.
Differences emerged at moderate strengths ($\epsilon \in \{0.01, 0.05, 0.1\}$), where \gls*{ato} consistently outperformed \gls*{ot} and \gls*{ate}.
For example, at $\epsilon=0.05$, the \gls*{prs} of \gls*{audioprotopnet} increased from 0.02 with \gls*{ot} to 0.09 with \gls*{ate} and 0.56 with \gls*{ato}.
At these moderate strengths, \gls*{audioprotopnet} with \gls*{ato} also achieved higher \gls*{prs} than ConvNeXt with \gls*{ato}. At $\epsilon=0.1$, their \gls*{prs} values were 0.35 and 0.09, respectively.
Even at $\epsilon=0.2$, where all models degraded severely, \gls*{audioprotopnet} with \gls*{ato} retained a \gls*{prs} of 0.13, highlighting the superior robustness conferred by \gls*{ato} against embedding-space attacks.
Subsequently, we evaluated robustness against untargeted output-space attacks. 
Figure~\ref{fig:perturbations_untargeted_os_attacks} illustrates typical examples of spectrograms perturbed by such attacks, alongside their clean counterparts and difference maps. 
The perturbations were often visually subtle, particularly in the more complex soundscape test instance, despite substantially affecting model performance.
Table~\ref{tab:robustness_untargeted_output_space} reports \gls*{prs} under output-space attacks.
At $\epsilon=0.001$, ConvNeXt and \gls*{audioprotopnet} achieved similar \gls*{prs} across all training approaches, ranging from 0.83 to 0.89.
As under embedding-space attacks, \gls*{ato} generally yielded the highest \gls*{prs} for both architectures, but output-space attacks degraded performance considerably more, producing generally lower \gls*{prs}, especially for stronger perturbations.
At $\epsilon\ge0.05$, \gls*{prs} fell below 0.01 for all models and training strategies, indicating substantial vulnerability.
At $\epsilon=0.01$, \gls*{audioprotopnet} with \gls*{ato} achieved 0.15 compared with 0.10 for ConvNeXt under the same training, still demonstrating the severe challenge of output-space attacks.
To probe how these attacks affected the models' internal states, we visualized the embeddings of clean and perturbed inputs from the ordinarily trained models, as shown in Figure~\ref{fig:embeddings_untargeted_attacks}. 
Using UMAP \cite{mcinnes2018umap}, we projected global average pooled embeddings of single-label instances from the five most frequent PER bird species after untargeted output-space and embedding-space \gls*{pgd} attacks at $\epsilon=0.05$.
Embedding-space attacks collapsed the clusters, whereas embeddings perturbed by output-space attacks retained some original structure despite equally devastating classification effects (Tables~\ref{tab:robustness_untargeted_embedding_space} and~\ref{tab:robustness_untargeted_output_space}).
\begin{table}[!ht]
  \caption{Untargeted embedding-space \gls*{prs} across $\epsilon$ for ConvNeXt and \gls*{audioprotopnet} under \gls*{ot}, \gls*{ato}, and \gls*{ate} (adversarial-training $\epsilon=0.1$). Best and second-best values across models and training strategies are bold and underlined, respectively, at each $\epsilon$.}
  \label{tab:robustness_untargeted_embedding_space}
  \centering
  \begin{tabular}{lccccc}
    \toprule
    \multicolumn{1}{c}{} 
      & \multicolumn{5}{c}{\textbf{Perturbation}} \\
    \cmidrule(lr){2-6}
    \textbf{Model} 
      & 0.001 & 0.01 & 0.05 & 0.1 & 0.2 \\
    \midrule
    ConvNeXt (OT)       
      & \underline{0.99} & 0.60 & 0.03 & 0.0 & 0.0 \\
    \midrule
    ConvNeXt (AT-E)     
      & \underline{0.99} & 0.58 & 0.12 & 0.04 & 0.01 \\
    \midrule
    ConvNeXt (AT-O)     
      & \textbf{1.0} & \underline{0.75} & \underline{0.27} & \underline{0.09} & \underline{0.02} \\
    \midrule
    AudioProtoPNet (OT) 
      & \textbf{1.0} & 0.74 & 0.02 & 0.0 & 0.0 \\
    \midrule
    AudioProtoPNet (AT-E)
      & \textbf{1.0} & 0.74 & 0.09 & 0.01 & 0.0 \\
    \midrule
    AudioProtoPNet (AT-O)
      & \textbf{1.0} & \textbf{0.89} & \textbf{0.56} & \textbf{0.35} & \textbf{0.13} \\
    \bottomrule
  \end{tabular}
\end{table}
\begin{table}[!ht]
  \caption{Untargeted output-space \gls*{prs} across $\epsilon$ for ConvNeXt and \gls*{audioprotopnet} under \gls*{ot}, \gls*{ato}, and \gls*{ate} (adversarial-training $\epsilon=0.1$). Best and second-best values are bold and underlined, respectively, at each $\epsilon$.}
  \label{tab:robustness_untargeted_output_space}
  \centering
  \begin{tabular}{lccccc}
    \toprule
    & \multicolumn{5}{c}{\textbf{Perturbation}} \\
    \cmidrule(lr){2-6}
    \textbf{Model} 
      & 0.001 & 0.01 & 0.05 & 0.1 & 0.2 \\
    \midrule
    ConvNeXt (OT)       
      & 0.83 & 0.02 & \textbf{0.0} & \textbf{0.0} & \textbf{0.0} \\
    \midrule
    ConvNeXt (AT-E)     
      & 0.85 & 0.04 & \textbf{0.0} & \textbf{0.0} & \textbf{0.0} \\
    \midrule
    ConvNeXt (AT-O)     
      & \textbf{0.89} & \underline{0.10} & \textbf{0.0} & \textbf{0.0} & \textbf{0.0} \\
    \midrule
    AudioProtoPNet (OT) 
      & 0.84 & 0.03 & \textbf{0.0} & \textbf{0.0} & \textbf{0.0} \\
    \midrule
    AudioProtoPNet (AT-E)
      & 0.86 & 0.09 & \textbf{0.0} & \textbf{0.0} & \textbf{0.0} \\
    \midrule
    AudioProtoPNet (AT-O)
      & \underline{0.88} & \textbf{0.15} & \textbf{0.0} & \textbf{0.0} & \textbf{0.0} \\
    \bottomrule
  \end{tabular}
\end{table}
This finding suggests that output-space attacks disrupt embedding space more structurally.
Overall, output-space attacks were considerably more potent and harder to defend against than embedding-space attacks across all tested models and training strategies, while \gls*{ato} protected against both attack types more effectively than \gls*{ate} or \gls*{ot}.
This suggests that \gls*{ato} fosters robustness not only in the final classifier layers but also deeper in the embedding extractor.
\Gls*{audioprotopnet} appeared to benefit more from \gls*{ato} than ConvNeXt, with higher clean-data performance and adversarial robustness under stronger perturbations.

\subsection{Stability of prototypes} 
\label{subsec:robustness_prototypes}
We investigated the stability of learned prototypes within \gls*{audioprotopnet} by evaluating its adversarial robustness against targeted embedding-space attacks. 
These attacks aimed to force an input's embedding map to resemble a specific, randomly selected target prototype within \gls*{audioprotopnet}. 
Table~\ref{tab:robustness_targeted_embedding_space} presents the \gls*{prs}, \gls*{drs}, and \gls*{tars} scores for these experiments across five perturbation strengths $\epsilon$; Figure~\ref{fig:targeted_attacks_scores} in Appendix~\ref{subsec:appendix_robustness_targeted_embedding_attacks} provides the corresponding visualization.
\begin{table}[!ht]
  \caption{Targeted embedding-space \gls*{prs}, \gls*{drs}, and \gls*{tars} across $\epsilon$ for \gls*{audioprotopnet} under \gls*{ot}, \gls*{ato}, and \gls*{ate} (adversarial-training $\epsilon=0.1$). \Gls*{ato} best preserved prototype stability. Best and second-best values per metric and $\epsilon$ are bold and underlined, respectively.}
  \label{tab:robustness_targeted_embedding_space}
  \centering
  \begin{tabular}{l l c c c c c}
    \toprule
    & & \multicolumn{5}{c}{\textbf{Perturbation}} \\
    \cmidrule(lr){3-7}
    \textbf{Model} & \textbf{Metric}
      & 0.001 & 0.01 & 0.05 & 0.1 & 0.2 \\
    \midrule
    \multirow{3}{*}{AudioProtoPNet (OT)}
      & PRS  & \textbf{1.0} & \textbf{0.90} & 0.38 & 0.18 & 0.07 \\
      & DRS  & \textbf{0.99} & 0.91            & 0.79            & 0.74 & 0.71 \\
      & TARS & \underline{0.99} & \textbf{0.91} & 0.49            & 0.27 & 0.12 \\
    \midrule
    \multirow{3}{*}{AudioProtoPNet (AT-E)}
      & PRS  & \textbf{1.0} & 0.86            & \underline{0.39} & \underline{0.26} & \underline{0.18} \\
      & DRS  & \textbf{0.99} & \underline{0.94} & \underline{0.85} & \underline{0.82} & \underline{0.81} \\
      & TARS & \textbf{1.0} & \underline{0.90}            & \underline{0.50} & \underline{0.37} & \underline{0.27} \\
    \midrule
    \multirow{3}{*}{AudioProtoPNet (AT-O)}
      & PRS  & \textbf{1.0} & \underline{0.87} & \textbf{0.57}    & \textbf{0.52}    & \textbf{0.43} \\
      & DRS  & \textbf{0.99} & \textbf{0.97}    & \textbf{0.94}    & \textbf{0.94}    & \textbf{0.92} \\
      & TARS & \textbf{1.0} & \textbf{0.91}    & \textbf{0.69}    & \textbf{0.65}    & \textbf{0.57} \\
    \bottomrule
  \end{tabular}
\end{table}
For ordinarily trained \gls*{audioprotopnet}, robustness declined rapidly at $\epsilon\ge0.05$: \gls*{prs} and \gls*{drs} fell below 0.38 and 0.79, respectively, yielding \gls*{tars} below 0.49.
Despite the substantial performance degradation reflected by the low \gls*{prs}, \gls*{drs} remained at least 0.71 even at $\epsilon=0.2$.
With \gls*{ate}, \gls*{tars} was comparable to \gls*{ot} for small perturbations ($\epsilon \in \{0.001,0.01,0.05\}$), but rose from 0.27 to 0.37 at $\epsilon=0.1$ and from 0.12 to 0.27 at $\epsilon=0.2$, providing moderate gains particularly under stronger attacks.
\Gls*{ato} followed a similar pattern with consistently larger prototype-stability gains.
It substantially increased \gls*{tars} for $\epsilon\ge0.05$, reaching 0.65 at $\epsilon=0.1$ and remaining at 0.57 at the maximum $\epsilon=0.2$, and therefore improved adversarial robustness more effectively than \gls*{ate}.
Because \gls*{tars} integrates classification robustness (\gls*{prs}) and embedding deformation (\gls*{drs}), high values indicate stable \gls*{audioprotopnet} explanations under adversarial input perturbations.
Thus, adversarial training, particularly \gls*{ato}, produced more robust prototype-based explanations than \gls*{ate} or \gls*{ot}.

\section{Discussion and Future Directions}
\label{sec:discussion}
For bird-sound classification, adversarial training improved \gls*{dl} models' clean-data performance, especially under strong distribution shifts, adversarial robustness under the evaluated \gls*{pgd} attacks, and prototype-explanation stability under targeted input perturbations.
These benefits require no architectural changes or inference overhead, making adversarial training a valuable complement to conventional audio augmentation in real-world systems.
However, TRADES-AWP increases training time and computation through adversarial-input generation and multiple optimization steps.
Across architectures, \gls*{ato} consistently provided greater overall benefits than \gls*{ate}, suggesting that direct decision-surface regularization is a potent augmentation for addressing distribution shifts.
Even under \gls*{ot}, \gls*{audioprotopnet} retained relatively high \gls*{drs} under targeted attacks, suggesting that despite impaired classification, its embedding maps did not easily align closely with an arbitrary target prototype.
Future work should explain \gls*{ato}'s broader gains and their interaction with architectural inductive biases, including \gls*{audioprotopnet}'s prototype mechanism.
The improved prototype-explanation stability motivates deeper study of adversarial training's explainability benefits through rigorous Quantus \cite{hedstrom2023quantus} or LATEC \cite{klein2024navigating} evaluation, human-centered practical-utility studies of stabilized explanations \cite{colin2022cannot}, and robustness-and-fidelity analyses of perturbation-based post-hoc methods such as SHAPIQ \cite{muschalik2024shapiq, lundberg2017unified}, LIME \cite{ribeiro2016should}, and occlusion sensitivity \cite{zeiler2014visualizing}.
Although \gls*{ato} performed best, \gls*{ate}'s gains remain relevant and could particularly benefit self-supervised pre-training of audio classifiers such as BirdMAE \cite{rauch2025can}.
Further work could adapt perturbation budgets dynamically and examine attacks beyond \gls*{fgsm} to balance clean-data performance and adversarial robustness.

Comprehensive field-deployment studies must assess responses to realistic, naturally occurring acoustic corruptions---reverberation, sensor discrepancies, and non-stationary background noise---and develop robustness against non-additive, domain-specific distortions, potentially by combining adversarial and tailored augmentation.
We studied two ConvNeXt-B-based bird-sound architectures: ConvNeXt-B, a strong domain baseline \cite{rauch2024birdset}, and \gls*{audioprotopnet}, which adds an interpretable prototype head and achieved state-of-the-art bird sound classification performance \cite{heinrich2025audioprotopnet}.
Future work should extend this analysis to other model families, including transformer-based and hybrid backbones, to determine how their inductive biases interact with adversarial training.
Moreover, although adversarial training is broadly applicable in principle, the observed benefits and optimal configurations may not transfer directly to industrial machinery-fault detection, environmental acoustic-scene classification, or other audio domains with different acoustic characteristics, data complexities, and objectives, potentially requiring tailored adversarial training.
These methods therefore require evaluation and adaptation across broader audio applications.
Together, these directions are pivotal to maturing adversarial training from a specialized technique into a standard, indispensable component for robust audio classifiers that reliably navigate complex real-world acoustic environments and their inherent distribution shifts.

\section{Conclusion}
\label{sec:conclusion}
For bird-sound classification under challenging train--test distribution shifts, adversarial training offered benefits beyond adversarial defense.
It substantially improved clean-data performance while improving adversarial robustness under the evaluated \gls*{pgd} attacks, supporting bioacoustic generalization.
Furthermore, integrating adversarial training with prototype-based models such as \gls*{audioprotopnet} improved the stability of the learned prototypes under targeted embedding-space attacks.
This integration also improved overall classification performance and model interpretability, the latter being critical for fostering user trust in applications where explainability is paramount.
Despite promising bioacoustic results, the direct applicability and optimal configurations of these methods may differ in other audio domains, including industrial machinery-fault detection and environmental acoustic-scene classification, because of their distinct acoustic characteristics, data complexities, and objectives.
These areas warrant dedicated future investigation, and we anticipate that our findings will encourage continued exploration and innovation in this research area, ultimately leading to more robust and reliable audio processing systems.
\PaperAcknowledgements

\section*{Code availability}
The code for the experiments and visualizations in this paper is available in an anonymous repository at \url{https://anonymous.4open.science/r/AttackAudioProtoPNet-75C8}.

\PaperAIDisclosure

\PaperBibliography

\appendix

\section{Background on Adversarial Attacks and Training}
\label{sec:appendix_adversarial_background}

Adversarial attacks are deliberate input manipulations designed to cause \gls*{ml} models to produce incorrect outputs. 
These attacks typically involve small, norm-bounded perturbations, yet capable of significantly degrading model performance \cite{xu2020adversarial}.
A comprehensive understanding of these attacks is therefore crucial for evaluating and enhancing the robustness of \gls*{ml} models.
Adversarial attacks are primarily categorized by the adversary's knowledge of the target model and the adversary's intended outcome \cite{xu2020adversarial}. 

\paragraph{Adversary knowledge.}
The first categorization criterion distinguishes between white-box, gray-box, and black-box approaches \cite{xu2020adversarial}.  
White-box attacks presume full access to the model, encompassing its architecture, parameters, and internal mechanisms. 
Such comprehensive knowledge enables direct gradient-based attacks and empirical evaluation under a white-box threat model.
Gray-box attacks operate with partial model information, for instance, access to intermediate outputs or confidence scores. 
Black-box attacks proceed with no internal knowledge, relying exclusively on observing the model's input-output behavior, rendering them more challenging yet often more aligned with realistic threat scenarios.
Throughout this work we assume a white-box adversary (see Section~\ref{subsec:attacks}).

\paragraph{Adversary objectives.}
The second categorization criterion is the adversary's objective. 
For classification tasks, output-space adversarial attacks are commonly categorized as untargeted or targeted.
Targeted attacks aim to compel the model to misclassify an input into a specific, attacker-chosen category. 
Untargeted attacks, conversely, seek only to induce any misclassification, without specifying the resultant incorrect class. 
However, this taxonomy is not directly suitable for tasks involving continuous outputs, such as the prediction of embeddings.
Following \citet{heinrich2024targeted}, we therefore distinguish between untargeted and targeted embedding-space attacks. 
Untargeted embedding-space attacks maximize the dissimilarity between the embeddings of clean and perturbed inputs, whereas targeted embedding-space attacks drive the embedding of a perturbed input toward a specified target vector.
In this work, we consider untargeted output-space attacks as well as untargeted and targeted embedding-space attacks.
This choice supports our evaluation of both generalization and adversarial robustness, and it enables the analysis of prototype stability in AudioProtoPNet.

\paragraph{Adversarial training.}
To defend against adversarial attacks, various strategies exist \cite{qiu2019review, xu2020adversarial, akhtar2021advances}, including methods that detect and remove manipulated data points at an early stage \cite{metzen2017detecting}. 
Another promising approach is to enhance model robustness by ensuring stability under small input perturbations \cite{szegedy2013intriguing}. 
In the case of output-space attacks, minor changes in the input should not lead to significant alterations in the model's classifications, whereas for embedding-space attacks, adversarial perturbations should have minimal impact on the model's predicted embeddings.
A prominent and effective defense mechanism is adversarial training, which directly improves the model's robustness to adversarial perturbations \cite{goodfellow2014explaining}. 
This training strategy augments the training dataset with adversarial inputs generated on-the-fly, effectively teaching the model to withstand such perturbations. 
At each training iteration, adversarial inputs are constructed based on the current model parameters, ensuring continuous adaptation to evolving adversarial threats. 
This process involves a minimax optimization where the model learns to minimize a loss function under worst-case input and parameter perturbations. 


\section{Datasets, preprocessing, and augmentation}
\label{sec:appendix_preprocessing}

Model evaluation uses all seven soundscape test datasets provided by BirdSet:
High Sierra Nevada (HSN) \cite{high_sieras_mary_clapp_2023_7525805},
NIPS4Bplus (NBP) \cite{morfi2019nips4bplus},
Colombia Costa Rica (NES) \cite{columbia_alvaro_vega_hidalgo_2023_7525349},
Amazon Basin (PER) \cite{amazon_basin_w_alexander_hopping_2022_7079124},
Sierra Nevada (SNE) \cite{sierra_nevada_stefan_kahl_2022_7050014},
Sapsucker Woods (SSW) \cite{sapsucker_stefan_kahl_2022_7079380},
and the Hawaiian Islands (UHH) \cite{hawaii_amanda_navine_2022_7078499}.
These datasets span diverse geographic regions (North America, South America, Oceania, and Europe) and cover 21 to 132 bird species.
Each test recording is divided into non-overlapping 5-second segments, which may contain bird vocalizations or represent no-call instances.

Following the BirdSet protocol, the number of training recordings per class is capped at 500, with no minimum threshold imposed.
The bambird event detection package \cite{michaud2023unsupervised} provides timestamps of detected vocalizations within each Xeno-Canto recording; for classes exceeding 500 recordings, exactly one event per recording is extracted to enforce this cap.
Each training instance is standardized to a duration of 5\,seconds, with shorter clips zero-padded \cite{denton2022improving, rauch2024birdset}.
Validation data comprise a random 20\% split of the curated training recordings at the recording level, following \citet{rauch2024birdset}.
During training, data augmentation techniques are applied at both the waveform and spectrogram level to improve generalization.
Waveform-level augmentations include time-shifting around detected events, background noise mixing from BirdVox-DCASE-20k \cite{lostanlen_2018_dcase}, colored noise injection, and gain adjustment.
Spectrogram-level augmentations include frequency and time masking \cite{park2019specaugment}.
To bridge the gap between the single-label focal training data and the multi-label soundscape evaluation setting, synthetic multi-label instances are generated via Multi-label Mixup \cite{zhang2017mixup, hendrycks2019augmix}, and no-call segments from BirdVox-DCASE-20k are injected as negative training instances.

Table~\ref{tab:preprocessing_augmentation_params} summarizes the preprocessing and data augmentation parameters used in this study.
All parameters follow the protocols established in \citet{rauch2024birdset} and \citet{heinrich2025audioprotopnet}.

\begin{table}[!ht]
  \caption{Preprocessing and data augmentation parameters. All parameters follow \citet{rauch2024birdset} and \citet{heinrich2025audioprotopnet}.}
  \label{tab:preprocessing_augmentation_params}
  \centering
  \begin{tabular}{@{}llc@{}}
    \toprule
    \textbf{Category} & \textbf{Parameter} & \textbf{Value} \\
    \midrule
    \multirow{6}{*}{Spectrogram}
      & Sample rate & 32\,kHz \\
      & Mel frequency bins & 256 \\
      & \gls*{fft} size & 2048 \\
      & Hop length & 256 \\
      & \gls*{stft} bins & 1025 \\
      & Z-score ($\mu$, $\sigma$) & $-13.369$, $13.162$ \\
    \midrule
    \multirow{4}{*}{\shortstack[l]{Waveform-level \\ augmentation}}
      & Time-shifting ($p$) & 1.0 \\
      & Background noise mixing ($p$) & 0.5 \\
      & Colored noise ($p$) & 0.2 \\
      & Gain adjustment ($p$) & 0.2 \\
    \midrule
    \multirow{2}{*}{\shortstack[l]{Spectrogram-level \\ augmentation}}
      & Frequency masking ($p$) & 0.5 \\
      & Time masking ($p$) & 0.3 \\
    \midrule
    \multirow{2}{*}{\shortstack[l]{Multi-label \\ augmentation}}
      & Multi-label Mixup ($p$, max instances) & 0.8, 3 \\
      & No-call injection ($p$) & 0.075 \\
    \midrule
    \multirow{2}{*}{Training data}
      & Max recordings per class & 500 \\
      & Instance duration & 5\,s \\
    \bottomrule
  \end{tabular}
\end{table}

\section{Training configuration}
\label{sec:appendix_training_config}

Table~\ref{tab:training_hyperparams} summarizes the model architecture and training hyperparameters used in this study.
The shared training configuration follows \citet{heinrich2025audioprotopnet} and uses AdamW \cite{loshchilov2017decoupled} with cosine annealing \cite{loshchilov2016sgdr}, while the adversarial-training configuration follows \citet{jin2023randomized} and \citet{wu2020adversarial}.
Both architectures comprise approximately 88 million parameters, with the exact count depending on the number of target classes.
For \gls*{audioprotopnet}, the higher prototype learning rate allows the prototypes to converge within the same training budget as the backbone and linear output layer \cite{heinrich2025audioprotopnet}.
Setting $\lambda=1.0$ weights the classification loss and robustness regularization term equally.
Each model configuration is trained five times with different random seeds, and the instance with the lowest validation loss is selected for evaluation.

\begin{appendixstricttable}
  \caption{Model architecture, optimization, adversarial-training, and model-selection specifications.}
  \label{tab:training_hyperparams}
  \centering
  \begin{tabular}{@{}llc@{}}
    \toprule
    \textbf{Category} & \textbf{Parameter} & \textbf{Value} \\
    \midrule
    \multirow{4}{*}{Architecture}
      & Backbone & ConvNeXt-B \\
      & Embedding dimensions ($H_z \times W_z \times D$) & $8 \times 19 \times 1024$ \\
      & Prototypes per class (\gls*{audioprotopnet}) & 20 \\
      & Parameters (approx.) & $\sim$88\,M \\
    \midrule
    \multirow{7}{*}{\shortstack[l]{Shared training}}
      & Optimizer & AdamW \\
      & Weight decay & $1 \times 10^{-4}$ \\
      & LR schedule & Cosine annealing (5\% warm-up) \\
      & Epochs & 20 \\
      & Batch size & 64 \\
      & Loss function & $\mathcal{L}_{asym}$ \cite{ridnik2021asymmetric} \\
      & GPU & NVIDIA A100-SXM4-80GB \\
    \midrule
    {\shortstack[l]{ConvNeXt \\ learning rates}}
      & Backbone + classifier & $1 \times 10^{-4}$ \\
    \midrule
    \multirow{2}{*}{\shortstack[l]{\gls*{audioprotopnet} \\ learning rates}}
      & Backbone + linear layer & $1 \times 10^{-4}$ \\
      & Prototype vectors & $1 \times 10^{-2}$ \\
    \midrule
    \multirow{5}{*}{\shortstack[l]{Adversarial \\ training (AT)}}
      & TRADES $\lambda$ & 1.0 \\
      & \gls*{awp} $\gamma$ & 0.005 \\
      & \gls*{awp} warm-up & 8 epochs \\
      & Weight randomization & $1 \times 10^{-4}$ \\
      & $\epsilon$ search range & $\{0.001, 0.01, 0.05, 0.1, 0.2\}$ \\
    \midrule
    \multirow{2}{*}{Evaluation}
      & Random seeds & 5 \\
      & Model selection & Lowest validation loss \\
    \bottomrule
  \end{tabular}
\end{appendixstricttable}

\section{Projected gradient descent attack optimization}
\label{sec:appendix_pgd_optimization}

All three evaluation attacks use the same projected sign-gradient update \cite{kurakin2018adversarial}:
\begin{equation}
  \mathbf{x}_{adv}^{(t+1)}
  =
  \operatorname{clip}_{\mathbf{x},\epsilon}
  \left\{
    \mathbf{x}_{adv}^{(t)}
    +
    \alpha\,\operatorname{sign}
    \left(
      \nabla_{\mathbf{x}_{adv}^{(t)}}
      \mathcal{J}\!\left(\mathbf{x}_{adv}^{(t)}\right)
    \right)
  \right\}.
  \label{eq:pgd_generic}
\end{equation}
Here, $\mathbf{x}_{adv}^{(0)}$ is initialized once at random within the $\ell_\infty$ ball of radius $\epsilon$ centered at $\mathbf{x}$, $\alpha$ is the step size, and $\operatorname{clip}_{\mathbf{x},\epsilon}$ projects each update back into this ball.
Thus, $\|\mathbf{x}_{adv}^{(t)}-\mathbf{x}\|_\infty\le\epsilon$ at every iteration.
The three attack objectives are
\begin{equation}
\begin{aligned}
  \mathcal{J}_{\mathrm{out}}(\mathbf{x}')
    &= \mathcal{L}\!\left(f_{\boldsymbol{\theta}}(\mathbf{x}'),\mathbf{y}\right),
    && \text{untargeted output-space},\\
  \mathcal{J}_{\mathrm{emb\text{-}u}}(\mathbf{x}')
    &= \mathcal{D}_{avg}\!\left(f_{\boldsymbol{\theta}_b}^{b}(\mathbf{x}'),\mathbf{z}\right),
    && \text{untargeted embedding-space},\\
  \mathcal{J}_{\mathrm{emb\text{-}t}}(\mathbf{x}')
    &= -\mathcal{D}_{min}\!\left(f_{\boldsymbol{\theta}_b}^{b}(\mathbf{x}'),\tilde{\mathbf{z}}_{adv}\right),
    && \text{targeted embedding-space}.
\end{aligned}
\label{eq:pgd_objective_mappings}
\end{equation}
Equation~\ref{eq:pgd_generic} therefore performs gradient ascent for both untargeted objectives; maximizing the negative distance in the targeted case is equivalent to minimizing $\mathcal{D}_{min}$.

\section{Formal metric definitions}
\label{sec:appendix_formal_metric_definitions}

\subsection{Class Mean Average Precision}
\label{subsec:appendix_cmap_definition}
Following \citet{rauch2025can}, we primarily assessed the performance of both ConvNeXt and \gls*{audioprotopnet} using the \gls*{cmap}.
The \gls*{cmap} is a suitable metric for multi-label audio classification scenarios, such as bird sound classification, because it comprehensively evaluates a model's ability to distinguish between multiple classes.
To formalize our evaluation, we first define the necessary notation for labels and classifications.
The true labels are represented as a binary matrix $\mathbf{Y} \in \{0,1\}^{N \times K}$, where $N$ is the total number of instances and $K$ is the number of classes.
Each element $y_{n,k} \in \{0,1\}$ of this matrix is defined as:
\begin{equation}
 y_{n,k} =
\begin{cases}
    1, & \text{if instance } n \text{ is labeled with class } k \\
    0, & \text{otherwise}
\end{cases}
\end{equation}
This element $y_{n,k}$ thus indicates whether class $k$ is present in instance $n$.
Similarly, the matrix of predicted confidence scores is denoted as $\mathbf{\hat{Y}} \in [0,1]^{N \times K}$, where each element $\hat{y}_{n,k} \in [0,1]$ corresponds to the model's predicted confidence that instance $n$ belongs to class $k$.
For clarity in subsequent formulas, we use $\mathbf{Y}_{:,k}$ and $\mathbf{\hat{Y}}_{:,k}$ to refer to the true labels and predicted confidence scores for a specific class $k$ across all $N$ instances, respectively.
Conversely, $\mathbf{Y}_{n,:}$ and $\mathbf{\hat{Y}}_{n,:}$ denote the true label vector and predicted confidence score vector for a specific instance $n$ across all $K$ classes.
The \gls*{cmap} is calculated by averaging the \gls*{ap} scores across all $K$ classes, providing an equitable measure irrespective of class frequency:
\begin{equation}
    \operatorname{cmAP} \left( \mathbf{\hat{Y}}, \mathbf{Y} \right) = \frac{1}{K} \sum_{k=1}^{K} \operatorname{AP}(\mathbf{\hat{Y}}_{:,k}, \mathbf{Y}_{:,k}).
\end{equation}
In this formulation, $\operatorname{AP}(\mathbf{\hat{Y}}_{:,k}, \mathbf{Y}_{:,k})$ denotes the \gls*{ap} for class $k$.
\gls*{ap} summarizes the precision-recall curve and reflects the quality of the model's ranking of instances for that particular class.
A high \gls*{cmap} score therefore implies that the model effectively identifies and distinguishes the vocalizations of individual bird species.
While \gls*{cmap} ensures equal weighting across classes regardless of their prevalence in the dataset \cite{kahl2023overview}, it is worth noting that its stability can be affected for classes with an extremely small number of positive instances \cite{denton2022improving}.

\subsection{Adversarial robustness evaluation metrics}
\label{subsec:appendix_robustness_metrics_definitions}

We assessed the empirical robustness of the two models under bounded input perturbations using \gls*{pgd} attacks.
These attacks comprised untargeted output-space attacks as well as both untargeted and targeted embedding-space attacks.
Specifically, we confined perturbations to the $\ell_\infty$-norm ball with maximum magnitudes of $\epsilon \in \{0.001, 0.01, 0.05, 0.1, 0.2\}$.
We executed the \gls*{pgd} attacks for $T = 10$ iterations, employing a step size of $\alpha = \frac{2\epsilon}{T}$.
For untargeted output-space and embedding-space attacks, we quantified adversarial robustness using the \glsentryfull{prs} \cite{heinrich2024targeted}.
In contrast, for targeted embedding-space attacks, we evaluated adversarial robustness using a comprehensive suite of three metrics.
These metrics, proposed by \citet{heinrich2024targeted}, were the \gls*{prs}, the \glsentryfull{drs}, and the \glsentryfull{tars}, which together facilitate a holistic assessment of adversarial robustness.
The \gls*{prs} measures the degradation in a model's performance when subjected to adversarial perturbations relative to its performance on clean inputs.
It is calculated using the \gls*{cmap} irrespective of whether the attack targets the model's output-space or its embedding-space.
Given the matrix of the model's predicted confidence scores on clean inputs $\mathbf{\hat{Y}}$, its classifications on the corresponding adversarial inputs $\mathbf{\hat{Y}}_{adv}$, and the true label matrix $\mathbf{Y}$, the \gls*{prs} is defined as:
\begin{equation}
\operatorname{PRS} \left( \mathbf{\hat{Y}}, \mathbf{\hat{Y}}_{adv}, \mathbf{Y} \right) = \min \left( \exp \left( 1 - \frac{\operatorname{cmAP} \left(\mathbf{\hat{Y}}, \mathbf{Y} \right)}{\operatorname{cmAP} \left(\mathbf{\hat{Y}}_{adv}, \mathbf{Y} \right) + \eta}\right), 1 \right).
\end{equation}
Here, $\eta = 1 \times 10^{-10}$ is a small constant included to prevent division by zero if the model's \gls*{cmap} on adversarial inputs is zero.
The \gls*{prs} ranges from 0, indicating severe performance degradation, to 1, signifying no loss in performance.
Adversarial robustness as measured by \gls*{prs} therefore decreases exponentially with the decline in \gls*{cmap} under attack.
In the context of targeted attacks, \gls*{prs} alone does not provide a complete picture of a model's adversarial robustness \cite{heinrich2024targeted}.
While an attacker might gauge success by the proximity of the adversarial embedding map $\mathbf{z}_{adv}$ to an adversarial target embedding vector $\tilde{\mathbf{z}}_{adv}$, this perspective does not encompass all potential vulnerabilities from the model's standpoint.
An attack might be deemed unsuccessful by the attacker if the model's prediction remains distant from the adversarial target.
However, if the attack still induces significant changes in the embedding map, a critical lack of robustness is evident.
Consequently, we also used \gls*{tars} \cite{heinrich2024targeted} to evaluate robustness against targeted embedding-space attacks, as it integrates \gls*{prs} with \gls*{drs}.
The \gls*{drs}, adapted from \citet{heinrich2024targeted}, quantifies the extent to which a targeted attack successfully distorts the model's embedding map towards a specified adversarial target embedding vector.
The dissimilarity between an original embedding map $\mathbf{z}$ and an adversarial target embedding vector $\tilde{\mathbf{z}}_{adv}$ is measured using the minimum spatial cosine distance $\mathcal{D}_{min}$ defined in Equation~\ref{eq:min_spatial_cosine_distance}.
Using this distance measure, the \gls*{drs} is formulated as:
\begin{equation}
    \operatorname{DRS} \left( \mathbf{z}, \mathbf{z}_{adv}, \tilde{\mathbf{z}}_{adv} \right) = \min \left( \exp \left( 1 - \frac{\mathcal{D}_{min} \left(\mathbf{z}, \tilde{\mathbf{z}}_{adv}\right)}{\mathcal{D}_{min} \left(\mathbf{z}_{adv}, \tilde{\mathbf{z}}_{adv}\right) + \eta}\right), 1 \right).
\end{equation}
A \gls*{drs} value approaching 0 indicates a successful deformation of the adversarial embedding map $\mathbf{z}_{adv}$ towards the adversarial target embedding vector $\tilde{\mathbf{z}}_{adv}$, whereas a value close to 1 implies minimal or no deformation.
The deformation robustness thus decreases exponentially as at least one of the regions of the adversarial embedding map converges towards the adversarial target embedding vector.
As neither \gls*{prs} nor \gls*{drs} in isolation sufficiently captures the nuances of adversarial robustness against targeted attacks, the \gls*{tars} \cite{heinrich2024targeted} is used.
\gls*{tars} integrates both \gls*{prs} and \gls*{drs} into a single score to provide a unified measure for targeted embedding-space attacks:
\begin{equation}
    \operatorname{TARS} _{\beta} = \left( 1 + \beta ^2 \right) \frac{\operatorname{PRS} \cdot \operatorname{DRS}}{\left( \beta ^2 \cdot \operatorname{PRS} \right) + \operatorname{DRS}}.
\end{equation}
The weighting parameter $\beta \in \mathbb{R}^+$ allows for adjustment of the relative importance of performance robustness (\gls*{prs}) versus deformation robustness (\gls*{drs}).
A higher $\beta$ value assigns greater weight to \gls*{drs}, while a lower value emphasizes \gls*{prs}.
In our evaluations, we set $\beta = 1$, thereby giving equal importance to \gls*{prs} and \gls*{drs}.
Critically, \gls*{tars} promotes a holistic view of robustness.
It yields high scores only when both \gls*{prs} and \gls*{drs} are concurrently high.
This characteristic makes \gls*{tars} particularly effective in identifying vulnerabilities from multiple perspectives.

\section{Distribution shift quantification}
\label{sec:appendix_distribution_shift}

We estimated the train--test distribution shift for each BirdSet subset by comparing class-specific distributions of Perch~v2 \cite{van2025perch} embeddings from the dedicated training data and the corresponding soundscape test data.
For each class, we computed a Fr\'echet distance between the empirical training and test embedding distributions and summarized the resulting class-level distances by their unweighted mean.

\paragraph{Setup.}
For each BirdSet subset, we loaded the dedicated training split and the corresponding test split using the BirdSet datamodule at 32\,kHz with instance peak normalization.
Training examples were 5\,s clips centered on precomputed event detections, whereas the soundscape recordings were divided into non-overlapping 5\,s clips.
We collected at most 5\,000 clips from each split and extracted one clip-level embedding vector per clip using Perch~v2.

\paragraph{Mean per-class Fr\'echet distance.}
For a fixed BirdSet subset $d$ and class $k$, let
$(\boldsymbol{\mu}_{\mathrm{train}}^{(d,k)},
\boldsymbol{\Sigma}_{\mathrm{train}}^{(d,k)})$
and
$(\boldsymbol{\mu}_{\mathrm{test}}^{(d,k)},
\boldsymbol{\Sigma}_{\mathrm{test}}^{(d,k)})$
denote the empirical mean and covariance of the corresponding Perch~v2 embeddings in the training and test splits, respectively.
Each covariance matrix was estimated using the
$(n_{s}^{(d,k)}-1)^{-1}$ normalization, where
$n_{s}^{(d,k)}$ is the number of retained clips for class $k$ in split
$s \in \{\mathrm{train},\mathrm{test}\}$.
For numerical stability, we regularized the covariance matrices as
\begin{equation}
  \widetilde{\boldsymbol{\Sigma}}_{s}^{(d,k)}
  =
  \boldsymbol{\Sigma}_{s}^{(d,k)}
  +
  10^{-6}\mathbf{I},
  \qquad
  s \in \{\mathrm{train},\mathrm{test}\},
\end{equation}
where $\mathbf{I}$ denotes the identity matrix of matching dimensionality.

The class-specific Fr\'echet distance was then computed as
\begin{equation}
\begin{aligned}
  d_{\mathrm{F}}^{(d,k)}
  ={}&
  \left\|
    \boldsymbol{\mu}_{\mathrm{train}}^{(d,k)}
    -
    \boldsymbol{\mu}_{\mathrm{test}}^{(d,k)}
  \right\|_2^2
  \\
  &+
  \operatorname{Tr}\!\Bigg(
    \widetilde{\boldsymbol{\Sigma}}_{\mathrm{train}}^{(d,k)}
    +
    \widetilde{\boldsymbol{\Sigma}}_{\mathrm{test}}^{(d,k)}
    -
    2
    \left[
      \left(
        \widetilde{\boldsymbol{\Sigma}}_{\mathrm{train}}^{(d,k)}
      \right)^{1/2}
      \widetilde{\boldsymbol{\Sigma}}_{\mathrm{test}}^{(d,k)}
      \left(
        \widetilde{\boldsymbol{\Sigma}}_{\mathrm{train}}^{(d,k)}
      \right)^{1/2}
    \right]^{1/2}
  \Bigg),
\end{aligned}
\end{equation}
where $\mathbf{A}^{1/2}$ denotes the principal positive-semidefinite matrix square root.

Only classes represented by at least 20 clips in both splits were included.
For each included class and split, we used at most 500 clips, selected with a fixed random seed.
Let $\mathcal{K}_d$ denote the set of eligible classes for subset $d$.
The dataset-level shift estimate was defined as the unweighted mean of the class-specific distances:
\begin{equation}
  D_{\mathrm{F}}(d)
  =
  \frac{1}{|\mathcal{K}_d|}
  \sum_{k \in \mathcal{K}_d}
  d_{\mathrm{F}}^{(d,k)}.
\end{equation}

\paragraph{Association with adversarial-training gains.}
The correlation analysis was conducted across the seven held-out soundscape test subsets
\begin{equation}
  \mathcal{D}_{\mathrm{test}}
  =
  \{
    \mathrm{PER},
    \mathrm{NES},
    \mathrm{UHH},
    \mathrm{HSN},
    \mathrm{NBP},
    \mathrm{SSW},
    \mathrm{SNE}
  \}.
\end{equation}
POW was excluded from this analysis because it was used for perturbation-strength selection rather than as a held-out test subset.

Let
\begin{equation}
  \mathcal{M}
  =
  \{
    \text{ConvNeXt},
    \text{AudioProtoPNet}
  \}
\end{equation}
denote the set of evaluated models, and let
$C_{d,m}^{(r)}$ denote the clean-data \gls*{cmap} obtained on subset $d$ by model $m$ under training regime $r$.
For each adversarial-training variant
$a \in \{\mathrm{AT\text{-}E},\mathrm{AT\text{-}O}\}$,
the model-specific change relative to ordinary training was
\begin{equation}
  \Delta_{d,m}^{(a)}
  =
  C_{d,m}^{(a)}
  -
  C_{d,m}^{(\mathrm{OT})}.
\end{equation}
We then calculated the equal-weight arithmetic mean of the two model-specific changes:
\begin{equation}
  \overline{\Delta}_{d}^{(a)}
  =
  \frac{1}{2}
  \sum_{m \in \mathcal{M}}
  \Delta_{d,m}^{(a)}.
\end{equation}
Thus, the two model-specific gains were averaged before ranking the datasets and calculating the correlation; the reported coefficient is not an average of two model-specific correlation coefficients.

Let $R_D(d)$ denote the rank of $D_{\mathrm{F}}(d)$ across
$\mathcal{D}_{\mathrm{test}}$, and let
$R_{\Delta}^{(a)}(d)$ denote the corresponding rank of
$\overline{\Delta}_{d}^{(a)}$.
Average ranks are used in the event of ties.
For each adversarial-training variant, Spearman's rank correlation was calculated as
\begin{equation}
  \rho_s^{(a)}
  =
  \operatorname{corr}\!\left(
    \left(R_D(d)\right)_{d \in \mathcal{D}_{\mathrm{test}}},
    \left(R_{\Delta}^{(a)}(d)\right)_{d \in \mathcal{D}_{\mathrm{test}}}
  \right),
\end{equation}
where $\operatorname{corr}(\cdot,\cdot)$ denotes the Pearson correlation between the two rank vectors.

\paragraph{Results.}
Table~\ref{tab:appendix_distribution_shift_results} reports the resulting dataset-level shift estimates.
Among the seven held-out soundscape test subsets, PER had the largest estimated shift, whereas HSN and UHH had the smallest estimates.
The Spearman rank correlation between the shift estimates and the equal-weight model-averaged \gls*{cmap} gains was
$\rho_s = 0.75$ for \gls*{ate} and
$\rho_s = 0.43$ for \gls*{ato}.
Both coefficients are positive, indicating that larger estimated train--test shifts tended to be accompanied by larger model-averaged gains from adversarial training.
The stronger correlation for \gls*{ate} reflects a closer correspondence between the cross-dataset ranking of the shift estimates and the ranking of the gains; it does not imply larger absolute performance gains than \gls*{ato}, which achieved the larger mean \gls*{cmap} improvements overall in Table~\ref{tab:clean_performance}.

\begin{tmlrappendixstricttable}
  \caption{
    Mean per-class Fr\'echet distance $D_{\mathrm{F}}(d)$ between the dedicated training data and the corresponding soundscape data in the Perch~v2 embedding space.
    Higher values indicate a larger estimated train--test distribution shift within this representation.
    POW, which was used for perturbation-strength selection, is shown for completeness but was excluded from the correlation analysis.
  }
  \label{tab:appendix_distribution_shift_results}
  \centering
  \begin{tabular}{lcccccccc}
    \toprule
    &
    \textbf{POW}
    &
    \textbf{PER}
    &
    \textbf{NES}
    &
    \textbf{UHH}
    &
    \textbf{HSN}
    &
    \textbf{NBP}
    &
    \textbf{SSW}
    &
    \textbf{SNE}
    \\
    \midrule
    $D_{\mathrm{F}}(d)$
    &
    7\,924
    &
    10\,925
    &
    9\,368
    &
    7\,692
    &
    7\,621
    &
    8\,724
    &
    7\,945
    &
    7\,974
    \\
    \bottomrule
  \end{tabular}
\end{tmlrappendixstricttable}

\section{Detailed results}
\label{sec:appendix_detailed_results}
This section provides supplementary information regarding the performance evaluation metrics employed in this research, alongside exhaustive tables detailing the results.

\subsection{Perturbation-strength ablation}
\label{subsec:appendix_perturbation_ablation}
Table~\ref{tab:clean_performance_POW} reports the complete clean-data performance ablation used to select the adversarial-training perturbation strength.
\begin{tmlrappendixstricttable}
  \centering
  \caption{\gls*{cmap} on the \gls*{pow} validation dataset for ConvNeXt and \gls*{audioprotopnet} under various training strategies. Performance is shown for ordinary training (\gls*{ot}) and both output-space adversarial training (\gls*{ato}) and embedding-space adversarial training (\gls*{ate}) at varying perturbation strengths $\epsilon$. Bold and underlined values indicate the best and second-best performance, respectively, for each training strategy.}
  \label{tab:clean_performance_POW}
  \begin{tabular}{@{}l c c c c c c@{}}
    \toprule
     &  
       & \multicolumn{5}{c}{\textbf{Perturbation (AT)}} \\
    \cmidrule(lr){3-7}
    \textbf{Model} 
      & \textbf{OT} 
      & \textbf{0.001} & \textbf{0.01} & \textbf{0.05} & \textbf{0.1} & \textbf{0.2} \\
    \midrule
    ConvNeXt (OT) 
      & 0.47   & -- & -- & -- & -- & -- \\
    \midrule
    ConvNeXt (AT-E) 
      &    --    & \textbf{0.54}\,{\scriptsize(+.07)} & \textbf{0.54}\,{\scriptsize(+.07)} & \textbf{0.54}\,{\scriptsize(+.07)} & \textbf{0.54}\,{\scriptsize(+.07)} & \underline{0.53}\,{\scriptsize(+.06)} \\
    \midrule
    ConvNeXt (AT-O) 
      &    --    & 0.53\,{\scriptsize(+.06)}  & 0.53\,{\scriptsize(+.06)}  & \underline{0.55}\,{\scriptsize(+.08)} & \textbf{0.56}\,{\scriptsize(+.09)} & 0.54\,{\scriptsize(+.07)}  \\
    \midrule
    AudioProtoPNet (OT)
      & 0.49   & -- & -- & -- & -- & -- \\
    \midrule
    AudioProtoPNet (AT-E) 
      &   --     & \underline{0.56}\,{\scriptsize(+.07)} & \textbf{0.57}\,{\scriptsize(+.08)} & \textbf{0.57}\,{\scriptsize(+.08)} & \textbf{0.57}\,{\scriptsize(+.08)} & \underline{0.56}\,{\scriptsize(+.07)} \\
    \midrule
    AudioProtoPNet (AT-O) 
      &    --    & 0.56\,{\scriptsize(+.07)}  & \underline{0.57}\,{\scriptsize(+.08)} & \textbf{0.59}\,{\scriptsize(+.10)} & \textbf{0.59}\,{\scriptsize(+.10)} & \textbf{0.59}\,{\scriptsize(+.10)} \\
    \bottomrule
  \end{tabular}
\end{tmlrappendixstricttable}

\subsection{Additional performance evaluation metrics}
\label{subsec:appendix_additional_metrics}
We further assessed the performance of both ConvNeXt and \gls*{audioprotopnet} using the \gls*{auroc} and Top-1 Accuracy, in addition to \gls*{cmap}. 
This approach followed the BirdSet evaluation protocol \cite{rauch2024birdset}. 
These metrics provide complementary insights into model capabilities for multi-label audio classification, particularly in the context of bird sound classification.
The \gls*{auroc} is a threshold-independent metric that evaluates a model's ability to distinguish between classes. 
It achieves this by measuring the area under the Receiver Operating Characteristic (ROC) curve. 
Formally, \gls*{auroc} represents the probability that a randomly chosen positive instance will be ranked higher than a randomly chosen negative instance. 
Its mathematical definition is \cite{van2024birds}:
\begin{equation}
        \operatorname{AUROC} \left( \mathbf{\hat{Y}}, \mathbf{Y} \right) = \frac{1}{K} \sum_{k=1}^{K} \left( \frac{1}{|\mathcal{P}_k| \cdot |\mathcal{N}_k|} \sum_{n \in \mathcal{P}_k} \sum_{m \in \mathcal{N}_k} \chi _{\left\{ \hat{y}_{n,k} > \hat{y}_{m,k} \right\}} \right).
\end{equation}
In this equation, $\mathcal{P}_k$ and $\mathcal{N}_k$ denote the sets of indices for positive and negative instances for class $k$, respectively. 
The term $\chi$ is the indicator function which equals one if the model assigns a higher score to the positive instance $n$ than to the negative instance $m$ for class $k$. 
A high \gls*{auroc} value indicates that the model confidently assigns higher scores to audio instances containing a specific bird species compared to instances without that species. 
This metric is also reliable when dealing with imbalanced class distributions \cite{hamer2023birb}.
Top-1 Accuracy measures if the class predicted with the highest confidence score is among the true labels for an audio instance \cite{rauch2024birdset}. 
We calculate it as:
\begin{equation}
        \text{T1-Acc} \left( \mathbf{\hat{Y}} , \mathbf{Y} \right) = \frac{1}{N} \sum_{n=1}^{N} \chi_{\left\{ y_{n, \hat{y}^{(\text{top})}_n} = 1 \right\}}.
\end{equation}
Here, $\hat{y}^{(\text{top})}_n = \argmax_{k} \hat{y}_{n,k}$ denotes the class predicted with the highest confidence for instance $n$, and the indicator equals one if this top-ranked class is a true label of instance $n$, that is, if $y_{n, \hat{y}^{(\text{top})}_n} = 1$.
While Top-1 Accuracy is straightforward and interpretable, showing if the model correctly identifies at least one relevant class, it does not fully capture the complexities of multi-label classification. 
For instance, in bird sound classification, this metric indicates the percentage of audio clips where the model correctly identifies at least one bird species. 
However, it does not assess whether all present species were correctly recognized. 
Moreover, Top-1 Accuracy can be disproportionately influenced by majority classes in datasets with imbalanced class distributions.

\subsection{Performance evaluation}
\label{subsec:appendix_overall_performance}
This section details the performance scores we obtained for the ConvNeXt and \gls*{audioprotopnet} models using various metrics. 
Table~\ref{tab:appendix_performance_scores_all_metrics} summarizes the \gls*{cmap}, \gls*{auroc}, and Top-1 Accuracy for these models under \gls*{ot}, \gls*{ate}, and \gls*{ato}. 
We present the results for the validation dataset (POW) and seven test datasets, in addition to the mean score across all test sets.
\begin{arxivcompactappendixtable}
  \caption{Detailed performance metrics (\gls*{cmap}, \gls*{auroc}, Top-1 Accuracy) for ConvNeXt and \gls*{audioprotopnet} on clean data. We compared \gls*{ot} with \gls*{ato} and \gls*{ate} using $\epsilon=0.1$. The table shows results for the POW validation set, seven test datasets, and the mean across these test datasets. Bold and underlined values indicate the best and second-best performance respectively for each dataset and metric. The key finding is that adversarial training, especially \gls*{ato}, generally enhanced performance for both models across the evaluated metrics and datasets.}
  \label{tab:appendix_performance_scores_all_metrics}
  \centering
  \resizebox{\textwidth}{!}{%
  \begin{tabular}{l l c c c c c c c c c}
    \toprule
    & & \multicolumn{8}{c}{\textbf{Dataset}} & \\
    \cmidrule(lr){3-10}
    \textbf{Model} & \textbf{Metric}
      & \textbf{POW} & \textbf{PER} & \textbf{NES} & \textbf{UHH}
      & \textbf{HSN} & \textbf{NBP} & \textbf{SSW} & \textbf{SNE}
      & \textbf{Score} \\
    \midrule
    \multirow{3}{*}{ConvNeXt (OT)}
      & cmAP   & 0.47 & 0.25 & 0.34 & 0.24 & 0.46 & \underline{0.68} & 0.37 & 0.31 & 0.38 \\
      & AUROC  & 0.85 & 0.76 & 0.88 & 0.78 & 0.89 & \underline{0.93} & 0.93 & 0.85 & 0.86 \\
      & T1-Acc & 0.76 & 0.50 & 0.49 & \textbf{0.38} & \textbf{0.61} & \textbf{0.71} & 0.53 & 0.68 & \underline{0.56} \\
    \midrule
    \multirow{3}{*}{ConvNeXt (AT-E)}
      & cmAP   & 0.54 & 0.29 & 0.37 & 0.23 & 0.43 & \underline{0.68} & \underline{0.41} & 0.32 & 0.39 \\
      & AUROC  & 0.89 & 0.78 & \textbf{0.91} & 0.79 & 0.89 & \underline{0.93} & \underline{0.94} & \underline{0.86} & 0.87 \\
      & T1-Acc & 0.84 & 0.56 & \underline{0.50} & 0.35 & \underline{0.56} & \underline{0.70} & \underline{0.59} & 0.68 & \underline{0.56} \\
    \midrule
    \multirow{3}{*}{ConvNeXt (AT-O)}
      & cmAP   & 0.56 & \underline{0.31} & \underline{0.39} & 0.23 & 0.47 & \textbf{0.69} & \textbf{0.44} & \underline{0.34} & \underline{0.41} \\
      & AUROC  & \underline{0.91} & \underline{0.80} & \textbf{0.91} & 0.79 & \textbf{0.91} & \textbf{0.94} & \textbf{0.95} & \textbf{0.87} & \underline{0.88} \\
      & T1-Acc & \textbf{0.89} & \textbf{0.59} & \textbf{0.54} & 0.33 & 0.53 & \textbf{0.71} & \textbf{0.61} & \textbf{0.70} & \textbf{0.57} \\
    \midrule
    \multirow{3}{*}{AudioProtoPNet (OT)}
      & cmAP   & 0.49 & 0.25 & 0.36 & \underline{0.25} & 0.46 & \underline{0.68} & 0.37 & 0.30 & 0.38 \\
      & AUROC  & 0.87 & 0.77 & 0.89 & \textbf{0.83} & 0.89 & \underline{0.93} & \underline{0.94} & 0.85 & 0.87 \\
      & T1-Acc & 0.76 & 0.49 & 0.48 & \underline{0.37} & \underline{0.56} & \underline{0.70} & 0.52 & 0.67 & 0.54 \\
    \midrule
    \multirow{3}{*}{AudioProtoPNet (AT-E)}
      & cmAP   & \underline{0.57} & 0.30 & 0.38 & \textbf{0.26} & \underline{0.48} & \textbf{0.69} & \underline{0.41} & 0.32 & \underline{0.41} \\
      & AUROC  & 0.90 & \underline{0.80} & \underline{0.90} & \underline{0.82} & \underline{0.90} & \textbf{0.94} & \underline{0.94} & \textbf{0.87} & \underline{0.88} \\
      & T1-Acc & 0.80 & 0.54 & \underline{0.50} & \textbf{0.38} & 0.50 & 0.69 & 0.56 & \underline{0.69} & 0.55 \\
    \midrule
    \multirow{3}{*}{AudioProtoPNet (AT-O)}
      & cmAP   & \textbf{0.59} & \textbf{0.32} & \textbf{0.40} & \underline{0.25} & \textbf{0.50} & \underline{0.68} & \textbf{0.44} & \textbf{0.35} & \textbf{0.42} \\
      & AUROC  & \textbf{0.92} & \textbf{0.82} & \underline{0.90} & \underline{0.82} & \underline{0.90} & \textbf{0.94} & \textbf{0.95} & \textbf{0.87} & \textbf{0.89} \\
      & T1-Acc & \underline{0.86} & \underline{0.57} & \textbf{0.54} & 0.36 & 0.52 & 0.69 & \underline{0.59} & \textbf{0.70} & \textbf{0.57} \\
    \bottomrule
  \end{tabular}
  }
\end{arxivcompactappendixtable}

\subsection{Adversarial robustness to untargeted attacks}
\label{subsec:appendix_robustness_untargeted_attacks}
We assessed the robustness of ConvNeXt and \gls*{audioprotopnet} against untargeted adversarial attacks using the \gls*{prs} metric. 
The subsequent tables provide these scores across eight datasets for five distinct perturbation strengths.

\subsubsection{Embedding-space attacks}
\label{subsubsec:appendix_embedding_untargeted_attacks}
Table~\ref{tab:appendix_prs_embedding_untargeted} shows the \gls*{prs} scores when models were subjected to untargeted embedding-space attacks. 
Adversarial training, particularly \gls*{ato} with \gls*{audioprotopnet}, substantially improved adversarial robustness against these attacks, especially at higher perturbation strengths.
\begin{table}[!ht]
  \caption{This table presents the \gls*{prs} for ConvNeXt and \gls*{audioprotopnet} against untargeted embedding-space attacks across various perturbation strengths $\epsilon$. We compared \gls*{ot} with \gls*{ato} and \gls*{ate} (trained with $\epsilon=0.1$). For each perturbation strength and dataset, bold and underlined values denote the best and second-best \gls*{prs} respectively, across models and training strategies. \gls*{audioprotopnet} with \gls*{ato} demonstrated superior robustness, maintaining higher \gls*{prs} values even at larger perturbation strengths.}
  \label{tab:appendix_prs_embedding_untargeted}
  \centering
  \resizebox{\textwidth}{!}{%
  \begin{tabular}{l l c c c c c c c c c}
    \toprule
    & & \multicolumn{8}{c}{\textbf{Dataset}} & \\
    \cmidrule(lr){3-10}
    \textbf{Model} & \textbf{Perturbation}
      & \textbf{POW} & \textbf{PER} & \textbf{NES} & \textbf{UHH}
      & \textbf{HSN} & \textbf{NBP} & \textbf{SSW} & \textbf{SNE}
      & \textbf{Score} \\
    \midrule
    \multirow{5}{*}{ConvNeXt (OT)}
      & 0.001 & \textbf{1.0} & \underline{0.99} & \underline{0.99} & \underline{0.99} & \textbf{1.0} & \textbf{1.0} & \underline{0.99} & \underline{0.99} & \underline{0.99} \\
      & 0.01  & 0.73 & 0.80 & 0.64 & 0.57 & 0.61 & 0.64 & 0.35 & 0.60 & 0.60 \\
      & 0.05  & 0.10 & 0.10 & 0.0 & 0.07 & 0.0 & 0.04 & 0.0 & 0.03 & 0.03 \\
      & 0.1   & 0.02 & 0.01 & 0.0 & 0.02 & 0.0 & 0.0 & \underline{0.0} & 0.0 & 0.0 \\
      & 0.2   & 0.0 & 0.0 & \underline{0.0} & 0.01 & 0.0 & 0.0 & \textbf{0.0} & 0.0 & 0.0 \\
    \midrule
    \multirow{5}{*}{ConvNeXt (AT-E)}
      & 0.001 & \underline{0.99} & \underline{0.99} & \textbf{1.0} & \textbf{1.0} & \underline{0.98} & \textbf{1.0} & \underline{0.99} & \underline{0.99} & \underline{0.99} \\
      & 0.01  & 0.64 & 0.74 & 0.65 & 0.64 & 0.52 & 0.62 & 0.38 & 0.49 & 0.58 \\
      & 0.05  & 0.09 & \underline{0.27} & 0.04 & 0.20 & 0.08 & 0.09 & 0.0 & 0.17 & 0.12 \\
      & 0.1   & 0.04 & \underline{0.13} & 0.0 & 0.07 & 0.01 & 0.03 & \underline{0.0} & \underline{0.07} & 0.04 \\
      & 0.2   & 0.01 & \underline{0.02} & \underline{0.0} & 0.04 & 0.0 & 0.0 & \textbf{0.0} & \underline{0.01} & 0.01 \\
    \midrule
    \multirow{5}{*}{ConvNeXt (AT-O)}
      & 0.001 & \textbf{1.0} & \textbf{1.0} & \textbf{1.0} & \textbf{1.0} & \textbf{1.0} & \textbf{1.0} & \textbf{1.0} & \underline{0.99} & \textbf{1.0} \\
      & 0.01  & \underline{0.82} & 0.73 & \underline{0.86} & 0.75 & 0.75 & \underline{0.73} & \underline{0.69} & \underline{0.71} & \underline{0.75} \\
      & 0.05  & \underline{0.35} & 0.22 & \underline{0.25} & \underline{0.57} & \underline{0.35} & \underline{0.18} & \underline{0.06} & \underline{0.23} & \underline{0.27} \\
      & 0.1   & \underline{0.13} & 0.05 & \underline{0.02} & \underline{0.31} & \underline{0.10} & \underline{0.05} & \underline{0.0} & \underline{0.07} & \underline{0.09} \\
      & 0.2   & \underline{0.03} & 0.01 & \underline{0.0} & \underline{0.11} & \underline{0.01} & \underline{0.01} & \textbf{0.0} & \underline{0.01} & \underline{0.02} \\
    \midrule
    \multirow{5}{*}{AudioProtoPNet (OT)}
      & 0.001 & \textbf{1.0} & \textbf{1.0} & \textbf{1.0} & \textbf{1.0} & \textbf{1.0} & \textbf{1.0} & \textbf{1.0} & \underline{0.99} & \textbf{1.0} \\
      & 0.01  & 0.80 & \underline{0.84} & 0.83 & 0.72 & \underline{0.78} & \underline{0.73} & 0.60 & 0.69 & 0.74 \\
      & 0.05  & 0.09 & 0.09 & 0.0 & 0.04 & 0.0 & 0.0 & 0.0 & 0.02 & 0.02 \\
      & 0.1   & 0.01 & 0.0 & 0.0 & 0.01 & 0.0 & 0.0 & \underline{0.0} & 0.0 & 0.0 \\
      & 0.2   & 0.0 & 0.0 & \underline{0.0} & 0.0 & 0.0 & 0.0 & \textbf{0.0} & 0.0 & 0.0 \\
    \midrule
    \multirow{5}{*}{AudioProtoPNet (AT-E)}
      & 0.001 & \textbf{1.0} & \textbf{1.0} & \textbf{1.0} & \textbf{1.0} & \textbf{1.0} & \textbf{1.0} & \textbf{1.0} & \textbf{1.0} & \textbf{1.0} \\
      & 0.01  & 0.76 & 0.83 & 0.80 & \underline{0.82} & 0.77 & \underline{0.73} & 0.53 & \underline{0.71} & 0.74 \\
      & 0.05  & 0.16 & 0.21 & 0.07 & 0.12 & 0.06 & 0.07 & 0.01 & 0.11 & 0.09 \\
      & 0.1   & 0.03 & 0.02 & 0.0 & 0.02 & 0.0 & 0.01 & \underline{0.0} & 0.01 & 0.01 \\
      & 0.2   & 0.01 & 0.0 & \underline{0.0} & 0.0 & 0.0 & 0.0 & \textbf{0.0} & 0.0 & 0.0 \\
    \midrule
    \multirow{5}{*}{AudioProtoPNet (AT-O)}
      & 0.001 & \textbf{1.0} & \textbf{1.0} & \textbf{1.0} & \textbf{1.0} & \textbf{1.0} & \textbf{1.0} & \textbf{1.0} & \textbf{1.0} & \textbf{1.0} \\
      & 0.01  & \textbf{0.94} & \textbf{0.92} & \textbf{0.96} & \textbf{0.88} & \textbf{0.87} & \textbf{0.85} & \textbf{0.87} & \textbf{0.91} & \textbf{0.89} \\
      & 0.05  & \textbf{0.66} & \textbf{0.70} & \textbf{0.72} & \textbf{0.73} & \textbf{0.55} & \textbf{0.52} & \textbf{0.24} & \textbf{0.49} & \textbf{0.56} \\
      & 0.1   & \textbf{0.41} & \textbf{0.50} & \textbf{0.50} & \textbf{0.51} & \textbf{0.36} & \textbf{0.24} & \textbf{0.04} & \textbf{0.29} & \textbf{0.35} \\
      & 0.2   & \textbf{0.17} & \textbf{0.21} & \textbf{0.15} & \textbf{0.21} & \textbf{0.14} & \textbf{0.10} & \textbf{0.0} & \textbf{0.13} & \textbf{0.13} \\
    \bottomrule
  \end{tabular}
  }
\end{table}

\subsubsection{Output-space attacks}
\label{subsubsec:appendix_output_untargeted_attacks}
To complement our evaluation on untargeted embedding-space attacks, Table~\ref{tab:appendix_prs_output_untargeted} details the \gls*{prs} scores for both ConvNeXt and \gls*{audioprotopnet} when we subjected them to untargeted output-space attacks. 
Similar to untargeted embedding-space attacks, \gls*{ato} yielded the greatest enhancement in adversarial robustness, although all models were more vulnerable to output-space attacks compared to embedding-space attacks.
\begin{tmlrappendixstricttable}
  \caption{This table shows the \gls*{prs} for ConvNeXt and \gls*{audioprotopnet} against untargeted output-space attacks at various perturbation strengths $\epsilon$. We compared \gls*{ot} with \gls*{ato} and \gls*{ate} (trained with $\epsilon=0.1$). Bold and underlined values indicate the best and second-best \gls*{prs} respectively, across models and training strategies. While \gls*{ato} offered some improvement, all models exhibited greater susceptibility to output-space attacks, with \gls*{prs} dropping rapidly as perturbation strength increased.}
  \label{tab:appendix_prs_output_untargeted}
  \centering
  \resizebox{\textwidth}{!}{%
  \begin{tabular}{l l c c c c c c c c c}
    \toprule
    & & \multicolumn{8}{c}{\textbf{Dataset}} & \\
    \cmidrule(lr){3-10}
    \textbf{Model} & \textbf{Perturbation}
      & \textbf{POW} & \textbf{PER} & \textbf{NES} & \textbf{UHH}
      & \textbf{HSN} & \textbf{NBP} & \textbf{SSW} & \textbf{SNE}
      & \textbf{Score} \\
    \midrule
    \multirow{5}{*}{ConvNeXt (OT)}
      & 0.001 & 0.89 & 0.87 & 0.81 & 0.79 & \underline{0.88} & 0.88 & 0.76 & \underline{0.84} & 0.83 \\
      & 0.01  & 0.09 & 0.09 & 0.0 & 0.01 & 0.01 & 0.04 & \underline{0.0} & 0.01 & 0.02 \\
      & 0.05  & 0.0 & \textbf{0.0} & \textbf{0.0} & \textbf{0.0} & \textbf{0.0} & \textbf{0.0} & \textbf{0.0} & \textbf{0.0} & \textbf{0.0} \\
      & 0.1   & \textbf{0.0} & \textbf{0.0} & \textbf{0.0} & \textbf{0.0} & \textbf{0.0} & \textbf{0.0} & \textbf{0.0} & \textbf{0.0} & \textbf{0.0} \\
      & 0.2   & \textbf{0.0} & \textbf{0.0} & \textbf{0.0} & \textbf{0.0} & \textbf{0.0} & \textbf{0.0} & \textbf{0.0} & \textbf{0.0} & \textbf{0.0} \\
    \midrule
    \multirow{5}{*}{ConvNeXt (AT-E)}
      & 0.001 & 0.89 & \textbf{0.90} & 0.85 & 0.87 & 0.84 & 0.89 & 0.79 & 0.81 & 0.85 \\
      & 0.01  & 0.09 & 0.09 & 0.01 & 0.10 & 0.02 & 0.05 & \underline{0.0} & 0.0 & 0.04 \\
      & 0.05  & 0.0 & \textbf{0.0} & \textbf{0.0} & \textbf{0.0} & \textbf{0.0} & \textbf{0.0} & \textbf{0.0} & \textbf{0.0} & \textbf{0.0} \\
      & 0.1   & \textbf{0.0} & \textbf{0.0} & \textbf{0.0} & \textbf{0.0} & \textbf{0.0} & \textbf{0.0} & \textbf{0.0} & \textbf{0.0} & \textbf{0.0} \\
      & 0.2   & \textbf{0.0} & \textbf{0.0} & \textbf{0.0} & \textbf{0.0} & \textbf{0.0} & \textbf{0.0} & \textbf{0.0} & \textbf{0.0} & \textbf{0.0} \\
    \midrule
    \multirow{5}{*}{ConvNeXt (AT-O)}
      & 0.001 & \textbf{0.94} & \textbf{0.90} & \textbf{0.88} & \textbf{0.91} & \textbf{0.93} & \textbf{0.92} & \textbf{0.85} & \textbf{0.85} & \textbf{0.89} \\
      & 0.01  & \underline{0.29} & 0.12 & 0.04 & 0.14 & 0.12 & \textbf{0.22} & \underline{0.0} & 0.05 & \underline{0.10} \\
      & 0.05  & \underline{0.02} & \textbf{0.0} & \textbf{0.0} & \textbf{0.0} & \textbf{0.0} & \textbf{0.0} & \textbf{0.0} & \textbf{0.0} & \textbf{0.0} \\
      & 0.1   & \textbf{0.0} & \textbf{0.0} & \textbf{0.0} & \textbf{0.0} & \textbf{0.0} & \textbf{0.0} & \textbf{0.0} & \textbf{0.0} & \textbf{0.0} \\
      & 0.2   & \textbf{0.0} & \textbf{0.0} & \textbf{0.0} & \textbf{0.0} & \textbf{0.0} & \textbf{0.0} & \textbf{0.0} & \textbf{0.0} & \textbf{0.0} \\
    \midrule
    \multirow{5}{*}{AudioProtoPNet (OT)}
      & 0.001 & 0.88 & 0.87 & 0.84 & 0.82 & 0.87 & 0.86 & 0.78 & 0.83 & 0.84 \\
      & 0.01  & 0.07 & 0.08 & 0.02 & 0.03 & 0.07 & 0.03 & \underline{0.0} & 0.01 & 0.03 \\
      & 0.05  & 0.0 & \textbf{0.0} & \textbf{0.0} & \textbf{0.0} & \textbf{0.0} & \textbf{0.0} & \textbf{0.0} & \textbf{0.0} & \textbf{0.0} \\
      & 0.1   & \textbf{0.0} & \textbf{0.0} & \textbf{0.0} & \textbf{0.0} & \textbf{0.0} & \textbf{0.0} & \textbf{0.0} & \textbf{0.0} & \textbf{0.0} \\
      & 0.2   & \textbf{0.0} & \textbf{0.0} & \textbf{0.0} & \textbf{0.0} & \textbf{0.0} & \textbf{0.0} & \textbf{0.0} & \textbf{0.0} & \textbf{0.0} \\
    \midrule
    \multirow{5}{*}{AudioProtoPNet (AT-E)}
      & 0.001 & 0.90 & \textbf{0.90} & \underline{0.86} & 0.87 & \underline{0.88} & 0.88 & \underline{0.80} & 0.81 & 0.86 \\
      & 0.01  & 0.18 & \underline{0.17} & \underline{0.07} & \underline{0.15} & \underline{0.14} & 0.07 & \underline{0.0} & \underline{0.06} & 0.09 \\
      & 0.05  & 0.0 & \textbf{0.0} & \textbf{0.0} & \textbf{0.0} & \textbf{0.0} & \textbf{0.0} & \textbf{0.0} & \textbf{0.0} & \textbf{0.0} \\
      & 0.1   & \textbf{0.0} & \textbf{0.0} & \textbf{0.0} & \textbf{0.0} & \textbf{0.0} & \textbf{0.0} & \textbf{0.0} & \textbf{0.0} & \textbf{0.0} \\
      & 0.2   & \textbf{0.0} & \textbf{0.0} & \textbf{0.0} & \textbf{0.0} & \textbf{0.0} & \textbf{0.0} & \textbf{0.0} & \textbf{0.0} & \textbf{0.0} \\
    \midrule
    \multirow{5}{*}{AudioProtoPNet (AT-O)}
      & 0.001 & \underline{0.91} & \underline{0.89} & \textbf{0.88} & \underline{0.89} & \textbf{0.93} & \underline{0.90} & \textbf{0.85} & \textbf{0.85} & \underline{0.88} \\
      & 0.01  & \textbf{0.32} & \textbf{0.23} & \textbf{0.12} & \textbf{0.18} & \textbf{0.25} & \underline{0.20} & \textbf{0.01} & \textbf{0.08} & \textbf{0.15} \\
      & 0.05  & \textbf{0.04} & \textbf{0.0} & \textbf{0.0} & \textbf{0.0} & \textbf{0.0} & \textbf{0.0} & \textbf{0.0} & \textbf{0.0} & \textbf{0.0} \\
      & 0.1   & \textbf{0.0} & \textbf{0.0} & \textbf{0.0} & \textbf{0.0} & \textbf{0.0} & \textbf{0.0} & \textbf{0.0} & \textbf{0.0} & \textbf{0.0} \\
      & 0.2   & \textbf{0.0} & \textbf{0.0} & \textbf{0.0} & \textbf{0.0} & \textbf{0.0} & \textbf{0.0} & \textbf{0.0} & \textbf{0.0} & \textbf{0.0} \\
    \bottomrule
  \end{tabular}
  }
\end{tmlrappendixstricttable}

\subsection{Stability of prototypes}
\label{subsec:appendix_robustness_targeted_embedding_attacks}
This subsection details our evaluation of the robustness of \gls*{audioprotopnet}'s prototypes when we subjected them to targeted embedding-space adversarial attacks. 
We present results for \gls*{prs}, \gls*{drs}, and \gls*{tars} to offer a comprehensive assessment of prototype stability under such conditions.
\begin{appendixstrictfigure}
    \centering
    \includegraphics[trim={0cm 0cm 0cm 0cm},clip,width=\textwidth]{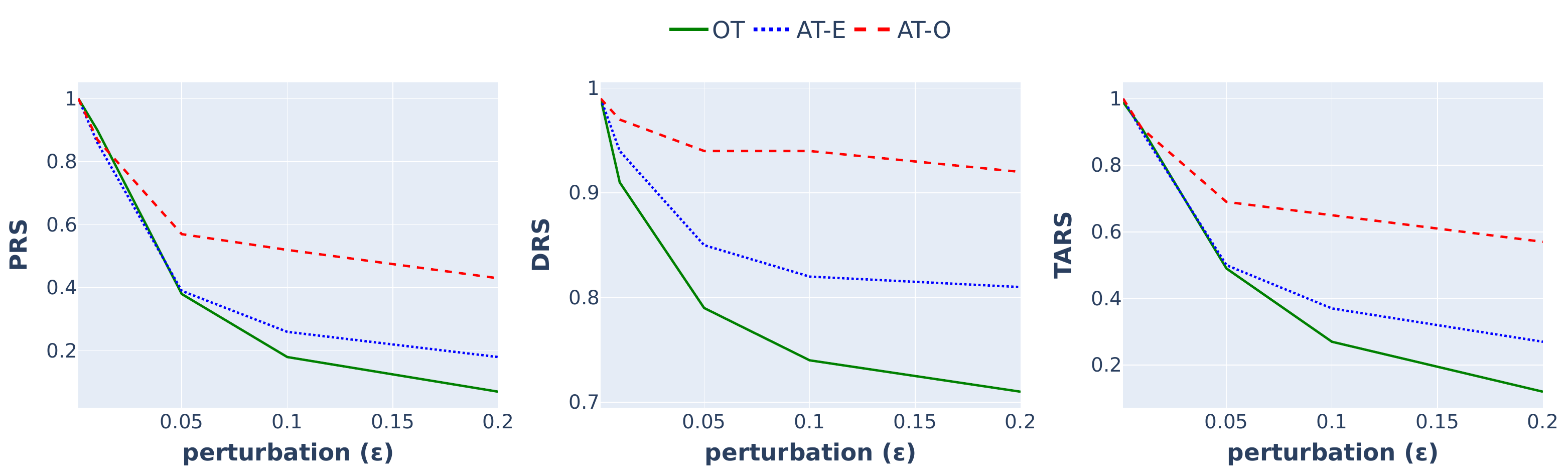}
    \caption{\gls*{prs}, \gls*{drs}, and \gls*{tars} scores for \gls*{audioprotopnet} against targeted embedding-space attacks, visually confirming that \gls*{ato} consistently yielded superior prototype stability across various perturbation strengths $\epsilon$. The figure compares \gls*{ot} with \gls*{ato} and \gls*{ate}, where adversarial training used a perturbation strength of $\epsilon=0.1$.}
    \label{fig:targeted_attacks_scores}
\end{appendixstrictfigure}

\subsubsection{PRS for targeted embedding-space attacks}
\label{subsubsec:appendix_prs_targeted_embedding_attacks}
Table~\ref{tab:appendix_prs_targeted_embedding_attacks} presents the \gls*{prs} scores we obtained for \gls*{audioprotopnet} under targeted embedding-space attacks. 
These scores reflected the model's ability to maintain correct classification despite these adversarial input perturbations. 
Adversarial training, particularly \gls*{ato}, consistently improved \gls*{prs} scores, indicating enhanced performance robustness.
\begin{appendixstricttable}
  \caption{This table shows the \gls*{prs} scores for \gls*{audioprotopnet} against targeted embedding-space attacks at various perturbation strengths $\epsilon$. We compared \gls*{ot} with \gls*{ato} and \gls*{ate} (trained with $\epsilon=0.1$). Bold and underlined values denote the best and second-best scores respectively across training strategies for each dataset and perturbation strength. \gls*{ato} demonstrated the most robust classification performance under these targeted attacks.} \label{tab:appendix_prs_targeted_embedding_attacks}
  \centering
  \resizebox{\textwidth}{!}{%
  \begin{tabular}{l l c c c c c c c c c}
    \toprule
    & & \multicolumn{8}{c}{\textbf{Dataset}} & \\
    \cmidrule(lr){3-10}
    \textbf{Model} & \textbf{Perturbation}
      & \textbf{POW} & \textbf{PER} & \textbf{NES} & \textbf{UHH}
      & \textbf{HSN} & \textbf{NBP} & \textbf{SSW} & \textbf{SNE}
      & \textbf{Score} \\
    \midrule
    \multirow{5}{*}{AudioProtoPNet (OT)}
      & 0.001 & \textbf{1.0} & \textbf{1.0} & \textbf{1.0} & \textbf{1.0} & \textbf{1.0} & \textbf{1.0} & \textbf{1.0} & \textbf{1.0} & \textbf{1.0} \\
      & 0.01  & \textbf{0.92} & \underline{0.92} & \textbf{0.94} & 0.90 & \textbf{0.91} & \textbf{0.94} & \textbf{0.82} & \textbf{0.90} & \textbf{0.90} \\
      & 0.05  & \underline{0.45} & 0.46 & \underline{0.32} & 0.48 & \underline{0.33} & \underline{0.62} & 0.08 & \underline{0.36} & 0.38 \\
      & 0.1   & \underline{0.27} & 0.23 & 0.09 & 0.33 & 0.08 & 0.38 & 0.01 & 0.17 & 0.18 \\
      & 0.2   & \underline{0.15} & 0.07 & 0.01 & 0.21 & 0.01 & 0.15 & 0.0 & 0.05 & 0.07 \\
    \midrule
    \multirow{5}{*}{AudioProtoPNet (AT-E)}
      & 0.001 & \textbf{1.0} & \textbf{1.0} & \textbf{1.0} & \textbf{1.0} & \textbf{1.0} & \textbf{1.0} & \textbf{1.0} & \underline{0.99} & \textbf{1.0} \\
      & 0.01  & \textbf{0.92} & \textbf{0.94} & \underline{0.90} & \underline{0.92} & \underline{0.86} & 0.90 & 0.73 & \underline{0.80} & 0.86 \\
      & 0.05  & 0.39 & \underline{0.50} & \underline{0.32} & \underline{0.53} & 0.27 & 0.59 & \underline{0.14} & 0.35 & \underline{0.39} \\
      & 0.1   & 0.25 & \underline{0.33} & \underline{0.20} & \underline{0.41} & \underline{0.09} & \underline{0.43} & \underline{0.11} & \underline{0.25} & \underline{0.26} \\
      & 0.2   & \underline{0.15} & \underline{0.21} & \underline{0.13} & \underline{0.32} & \underline{0.04} & \underline{0.29} & \underline{0.08} & \underline{0.16} & \underline{0.18} \\
    \midrule
    \multirow{5}{*}{AudioProtoPNet (AT-O)}
      & 0.001 & \textbf{1.0} & \textbf{1.0} & \textbf{1.0} & \textbf{1.0} & \textbf{1.0} & \textbf{1.0} & \textbf{1.0} & \textbf{1.0} & \textbf{1.0} \\
      & 0.01  & \underline{0.89} & 0.90 & \underline{0.90} & \textbf{0.94} & \textbf{0.91} & \underline{0.92} & \underline{0.76} & 0.79 & \underline{0.87} \\
      & 0.05  & \textbf{0.60} & \textbf{0.71} & \textbf{0.65} & \textbf{0.64} & \textbf{0.43} & \textbf{0.73} & \textbf{0.33} & \textbf{0.50} & \textbf{0.57} \\
      & 0.1   & \textbf{0.56} & \textbf{0.73} & \textbf{0.60} & \textbf{0.50} & \textbf{0.26} & \textbf{0.68} & \textbf{0.36} & \textbf{0.53} & \textbf{0.52} \\
      & 0.2   & \textbf{0.48} & \textbf{0.63} & \textbf{0.45} & \textbf{0.42} & \textbf{0.19} & \textbf{0.64} & \textbf{0.25} & \textbf{0.42} & \textbf{0.43} \\
    \bottomrule
  \end{tabular}
  }
\end{appendixstricttable}

\subsubsection{DRS for targeted embedding-space attacks}
\label{subsubsec:appendix_drs_targeted_attacks}
To further quantify the stability of prototype-based explanations under targeted embedding-space attacks, we employed the \gls*{drs} metric. 
Table~\ref{tab:appendix_drs_targeted_attacks} presents the \gls*{drs} scores for \gls*{audioprotopnet}. 
These scores measured the robustness of the model's embeddings against adversarial input perturbations designed to shift them towards randomly chosen prototypes of \gls*{audioprotopnet}. 
Higher \gls*{drs} scores, particularly evident with \gls*{ato}, indicated that the embedding maps exhibited greater robustness to such targeted deformations.
\begin{appendixcompactstricttable}
  \caption{This table presents the \gls*{drs} scores for \gls*{audioprotopnet} against targeted embedding-space attacks at various perturbation strengths $\epsilon$. We compared \gls*{ot} with \gls*{ato} and \gls*{ate} (trained with $\epsilon=0.1$). Bold and underlined values indicate the best and second-best scores respectively across training strategies for each dataset and perturbation strength. The key observation is that \gls*{ato} enhanced the deformation robustness of embedding maps, resulting in consistently higher \gls*{drs} scores.}
  \label{tab:appendix_drs_targeted_attacks}
  \centering
  \resizebox{\textwidth}{!}{%
  \begin{tabular}{l l c c c c c c c c c}
    \toprule
    & & \multicolumn{8}{c}{\textbf{Dataset}} & \\
    \cmidrule(lr){3-10}
    \textbf{Model} & \textbf{Perturbation}
      & \textbf{POW} & \textbf{PER} & \textbf{NES} & \textbf{UHH}
      & \textbf{HSN} & \textbf{NBP} & \textbf{SSW} & \textbf{SNE}
      & \textbf{Score} \\
    \midrule
    \multirow{5}{*}{AudioProtoPNet (OT)}
      & 0.001 & \textbf{0.99} & \underline{0.99} & \textbf{0.99} & \textbf{0.99} & \textbf{0.99} & \textbf{0.99} & \textbf{0.99} & \textbf{0.99} & \textbf{0.99} \\
      & 0.01  & 0.91 & 0.93 & 0.91 & 0.90 & 0.92 & 0.89 & 0.92 & 0.91 & 0.91 \\
      & 0.05  & 0.76 & 0.78 & 0.78 & 0.77 & 0.81 & 0.78 & 0.80 & 0.79 & 0.79 \\
      & 0.1   & 0.71 & 0.73 & 0.72 & 0.74 & 0.77 & 0.75 & 0.74 & 0.74 & 0.74 \\
      & 0.2   & 0.69 & 0.70 & 0.69 & 0.73 & 0.75 & 0.72 & 0.70 & 0.70 & 0.71 \\
    \midrule
    \multirow{5}{*}{AudioProtoPNet (AT-E)}
      & 0.001 & \textbf{0.99} & \textbf{1.0}  & \textbf{0.99} & \textbf{0.99} & \textbf{0.99} & \textbf{0.99} & \textbf{0.99} & \textbf{0.99} & \textbf{0.99} \\
      & 0.01  & \underline{0.94} & \underline{0.96} & \underline{0.94} & \underline{0.94} & \underline{0.93} & \underline{0.92} & \underline{0.94} & \underline{0.94} & \underline{0.94} \\
      & 0.05  & \underline{0.82} & \underline{0.86} & \underline{0.84} & \underline{0.84} & \underline{0.85} & \underline{0.84} & \underline{0.85} & \underline{0.85} & \underline{0.85} \\
      & 0.1   & \underline{0.77} & \underline{0.82} & \underline{0.81} & \underline{0.81} & \underline{0.82} & \underline{0.81} & \underline{0.83} & \underline{0.83} & \underline{0.82} \\
      & 0.2   & \underline{0.74} & \underline{0.80} & \underline{0.80} & \underline{0.79} & \underline{0.82} & \underline{0.80} & \underline{0.82} & \underline{0.81} & \underline{0.81} \\
    \midrule
    \multirow{5}{*}{AudioProtoPNet (AT-O)}
      & 0.001 & \textbf{0.99} & \underline{0.99} & \textbf{0.99} & \textbf{0.99} & \textbf{0.99} & \textbf{0.99} & \textbf{0.99} & \textbf{0.99} & \textbf{0.99} \\
      & 0.01  & \textbf{0.97} & \textbf{0.98} & \textbf{0.97} & \textbf{0.95} & \textbf{0.95} & \textbf{0.96} & \textbf{0.97} & \textbf{0.98} & \textbf{0.97} \\
      & 0.05  & \textbf{0.94} & \textbf{0.96} & \textbf{0.95} & \textbf{0.89} & \textbf{0.94} & \textbf{0.94} & \textbf{0.95} & \textbf{0.95} & \textbf{0.94} \\
      & 0.1   & \textbf{0.93} & \textbf{0.96} & \textbf{0.95} & \textbf{0.87} & \textbf{0.92} & \textbf{0.93} & \textbf{0.96} & \textbf{0.96} & \textbf{0.94} \\
      & 0.2   & \textbf{0.91} & \textbf{0.95} & \textbf{0.94} & \textbf{0.86} & \textbf{0.90} & \textbf{0.92} & \textbf{0.95} & \textbf{0.94} & \textbf{0.92} \\
    \bottomrule
  \end{tabular}
  }
\end{appendixcompactstricttable}

\subsubsection{TARS for targeted embedding-space attacks}
\label{subsubsec:appendix_tars_targeted_embedding_attacks}
Finally, we assessed the total adversarial robustness of \gls*{audioprotopnet} using the \gls*{tars} metric, which integrates performance robustness and deformation robustness. 
Table~\ref{tab:appendix_tars_targeted_embedding_attacks} details these comprehensive results. 
Our findings indicated that \gls*{audioprotopnet} models trained with \gls*{ato} achieved the highest \gls*{tars} scores, demonstrating superior overall robustness against targeted embedding-space attacks.
\begin{appendixcompactstricttable}
  \caption{This table shows the \gls*{tars} scores for \gls*{audioprotopnet} against targeted embedding-space attacks at various perturbation strengths $\epsilon$. We compared \gls*{ot} with \gls*{ato} and \gls*{ate} (trained with $\epsilon=0.1$). Bold and underlined values indicate the best and second-best scores respectively across training strategies for each dataset and perturbation strength. \gls*{ato} yielded the best overall robustness, achieving the highest \gls*{tars} scores by improving both performance and deformation robustness.} 
  \label{tab:appendix_tars_targeted_embedding_attacks}
  \centering
  \resizebox{\textwidth}{!}{%
  \begin{tabular}{l l c c c c c c c c c}
    \toprule
    & & \multicolumn{8}{c}{\textbf{Dataset}} & \\
    \cmidrule(lr){3-10}
    \textbf{Model} & \textbf{Perturbation}
      & \textbf{POW} & \textbf{PER} & \textbf{NES} & \textbf{UHH}
      & \textbf{HSN} & \textbf{NBP} & \textbf{SSW} & \textbf{SNE}
      & \textbf{Score} \\
    \midrule
    \multirow{5}{*}{AudioProtoPNet (OT)}
      & 0.001 & \underline{0.99} & \textbf{1.0}  & \textbf{1.0}  & \underline{0.99} & \underline{0.99} & \underline{0.99} & \underline{0.99} & \textbf{0.99} & \underline{0.99} \\
      & 0.01  & \underline{0.91} & 0.93 & \textbf{0.93} & 0.90 & \underline{0.91} & \underline{0.91} & \textbf{0.87} & \textbf{0.91} & \textbf{0.91} \\
      & 0.05  & \underline{0.57} & 0.58 & \underline{0.46} & 0.59 & \underline{0.46} & \underline{0.69} & 0.15 & \underline{0.49} & 0.49 \\
      & 0.1   & \underline{0.40} & 0.34 & 0.16 & 0.45 & 0.14 & 0.50 & 0.01 & 0.27 & 0.27 \\
      & 0.2   & 0.24 & 0.13 & 0.02 & 0.33 & 0.01 & 0.24 & 0.0  & 0.10 & 0.12 \\
    \midrule
    \multirow{5}{*}{AudioProtoPNet (AT-E)}
      & 0.001 & \textbf{1.0} & \textbf{1.0} & \textbf{1.0} & \textbf{1.0} & \underline{0.99} & \underline{0.99} & \textbf{1.0} & \textbf{0.99} & \textbf{1.0} \\
      & 0.01  & \textbf{0.93} & \textbf{0.95} & \underline{0.92} & \underline{0.93} & 0.89 & \underline{0.91} & 0.82 & 0.86 & \underline{0.90} \\
      & 0.05  & 0.53 & \underline{0.63} & 0.45 & \underline{0.65} & 0.39 & \underline{0.69} & \underline{0.23} & \underline{0.49} & \underline{0.50} \\
      & 0.1   & 0.38 & \underline{0.47} & \underline{0.29} & \underline{0.54} & \underline{0.16} & \underline{0.56} & \underline{0.16} & \underline{0.38} & \underline{0.37} \\
      & 0.2   & \underline{0.25} & \underline{0.33} & \underline{0.19} & \underline{0.45} & \underline{0.07} & \underline{0.43} & \underline{0.12} & \underline{0.27} & \underline{0.27} \\
    \midrule
    \multirow{5}{*}{AudioProtoPNet (AT-O)}
      & 0.001 & \textbf{1.0} & \textbf{1.0} & \textbf{1.0} & \textbf{1.0} & \textbf{1.0} & \textbf{1.0} & \textbf{1.0} & \textbf{0.99} & \textbf{1.0} \\
      & 0.01  & \textbf{0.93} & \underline{0.94} & \textbf{0.93} & \textbf{0.94} & \textbf{0.93} & \textbf{0.94} & \underline{0.85} & \underline{0.87} & \textbf{0.91} \\
      & 0.05  & \textbf{0.73} & \textbf{0.81} & \textbf{0.77} & \textbf{0.74} & \textbf{0.58} & \textbf{0.82} & \textbf{0.49} & \textbf{0.65} & \textbf{0.69} \\
      & 0.1   & \textbf{0.70} & \textbf{0.82} & \textbf{0.73} & \textbf{0.63} & \textbf{0.41} & \textbf{0.78} & \textbf{0.52} & \textbf{0.68} & \textbf{0.65} \\
      & 0.2   & \textbf{0.63} & \textbf{0.75} & \textbf{0.61} & \textbf{0.56} & \textbf{0.31} & \textbf{0.76} & \textbf{0.40} & \textbf{0.58} & \textbf{0.57} \\
    \bottomrule
  \end{tabular}
  }
\end{appendixcompactstricttable}

\end{document}